\documentclass[10pt,twocolumn,letterpaper]{article}

\usepackage{cvpr}              %
\usepackage[accsupp]{axessibility}

\definecolor{eccvblue}{rgb}{0.12,0.49,0.85}
\definecolor{lightcoral}{rgb}{0.94, 0.5, 0.5}
\definecolor{fuzzywuzzy}{rgb}{0.8, 0.4, 0.4}
\definecolor{frenchrose}{rgb}{0.96, 0.29, 0.54}
\definecolor{bronze}{rgb}{0.8, 0.5, 0.2}
\definecolor{cyann}{rgb}{0.0, 0.72, 0.92}
\definecolor{lightthulianpink}{rgb}{0.9, 0.56, 0.67}
\definecolor{colorone}{rgb}{0.77, 0.49, 0.97}
\definecolor{deepsaffron}{rgb}{1.0, 0.6, 0.2}
\definecolor{deepskyblue}{rgb}{0.0, 0.75, 1.0}
\definecolor{flame}{rgb}{0.89, 0.35, 0.13}
\definecolor{forestgreen}{rgb}{0.13, 0.55, 0.13}

\definecolor{cvprblue}{rgb}{0.21,0.49,0.74}
\usepackage[pagebackref,breaklinks,colorlinks,allcolors=cvprblue]{hyperref}
\usepackage{multirow} 
\usepackage{makecell}
\usepackage{booktabs}
\usepackage{marvosym} %
\usepackage{hyperref}

\def\confName{CVPR}
\def\confYear{2025}

\newcommand{\bs}{\mathbf{s}}

\newcommand{\bv}{\mathbf{v}}

\newcommand{\bp}{\mathbf{p}}

\newcommand{\bP}{\mathbf{P}}

\newcommand{\bff}{\mathbf{f}}
\newcommand{\bF}{\mathbf{F}}

\newcommand{\bc}{\mathbf{c}}

\newcommand{\bd}{\mathbf{d}}

\newcommand{\bmu}{\boldsymbol{\mu}}
\newcommand{\bSigma}{\boldsymbol{\Sigma}}

\newcommand{\balpha}{\boldsymbol{\alpha}}
\newcommand{\nR}{\mathbb{R}}

\newcommand{\cP}{\mathcal{P}}

\makeatletter
\DeclareRobustCommand\onedot{\futurelet\@let@token\@onedot}
\def\@onedot{\ifx\@let@token.\else.\null\fi\xspace}
\def\eg{e.g\onedot}

\def\wrt{wrt\onedot}

\def\Fig{Fig\onedot}   
\makeatother

\newcommand{\figref}[1]{\Fig~\ref{#1}}
\newcommand{\secref}[1]{Section~\ref{#1}}

\renewcommand{\eqref}[1]{Eq.~\ref{#1}}

\newcommand{\boldparagraph}[1]{\vspace{0.2cm}\noindent{\bf #1:} }

\newif\ifcomment
\commenttrue
\ifcomment
	\newcommand{\ag}[1]{ \noindent {\color{red} {\bf Andreas:} {#1}} }
	\newcommand{\ms}[1]{ \noindent {\color{orange} {\bf Sheng:} {#1}} }

\else
	\newcommand{\ag}[1]{}
	\newcommand{\yl}[1]{}
    \newcommand{\ms}[1]{}
\fi

\newcommand{\method}{EVolSplat}

\makeatletter
\def\blfootnote{\gdef\@thefnmark{}\@footnotetext}
\makeatother

\title{EVolSplat: Efficient Volume-based Gaussian Splatting for Urban View Synthesis}

\author{
Sheng Miao$^{1}$,
Jiaxin Huang$^{1}$,
Dongfeng Bai$^{2}$,
Xu Yan$^{2}$,
Hongyu Zhou$^{1}$,
Yue Wang$^{1}$,
\\
Bingbing Liu$^{2}$
Andreas Geiger$^{3,4}$,
Yiyi Liao$^{1}\textsuperscript{\Letter}$
\vspace{0.2cm}
\\
{\normalsize $^{1}$ Zhejiang University \quad $^{2}$ Huawei Noah's Ark Lab}
{\normalsize \quad $^{3}$ University of Tübingen \quad }
{\normalsize $^{4}$ Tübingen AI Center}
\\
\small{Project Page: \url{https://xdimlab.github.io/EVolSplat/}}
}

\begin{document}

\twocolumn[{%
\renewcommand\twocolumn[1][]{#1}%
\maketitle
\vspace{-1.4cm}
\begin{center}
    \centering
    \captionsetup{type=figure}
    \includegraphics[width=\textwidth]{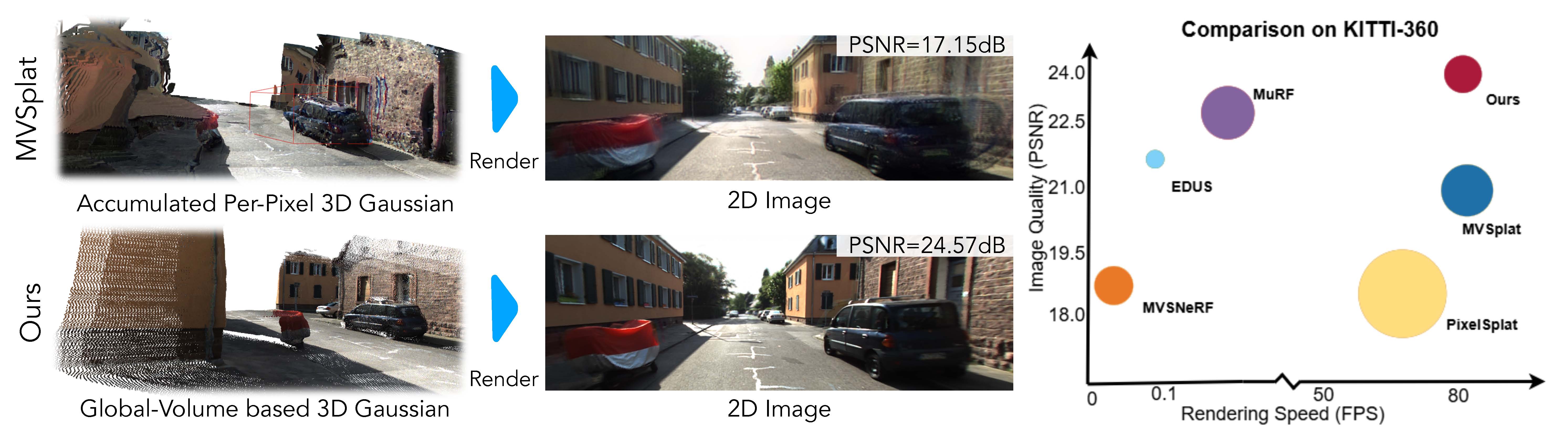}
    \captionof{figure}{\textbf{Illustration.} Left: Existing feed-forward 3DGS methods (e.g., MVSplat) predict per-pixel Gaussians with local cost volumes. When accumulating Gaussians from multiple local volumes in global coordinates, we observe inconsistencies in the accumulated Gaussians (\eg car in the figure), leading to ghost artifacts in the rendering. In contrast,~\method~predicts 3DGS using a global volume, improving consistency and rendering quality. Right: Our method achieves real-time rendering while maintaining high NVS rendering quality on novel street scenes with lower memory consumption. The circle size indicates memory consumption during inference. }
    \label{fig:teaser}
\end{center}%

}]

\renewcommand{\thefootnote}{}
\footnotetext[2]{$\textsuperscript{\Letter}$ Corresponding author.}
\renewcommand{\thefootnote}{\arabic{footnote}}

\begin{abstract}
Novel view synthesis of urban scenes is essential for autonomous driving-related applications. 
Existing NeRF and 3DGS-based methods show promising results in achieving photorealistic renderings but require slow, per-scene optimization. We introduce EVolSplat, an efficient 3D Gaussian Splatting model for urban scenes that works in a feed-forward manner. Unlike existing feed-forward, pixel-aligned 3DGS methods, which often suffer from issues like multi-view inconsistencies and duplicated content, our approach predicts 3D Gaussians across multiple frames within a unified volume using a 3D convolutional network. This is achieved by initializing 3D Gaussians with noisy depth predictions, and then refining their geometric properties in 3D space and predicting color based on 2D textures. Our model also handles distant views and the sky with a flexible hemisphere background model. This enables us to perform fast, feed-forward reconstruction while achieving real-time rendering. Experimental evaluations on the KITTI-360 and Waymo datasets show that our method achieves state-of-the-art quality compared to existing feed-forward 3DGS- and NeRF-based methods.
\end{abstract}

\vspace{-0.5cm}
\section{Introduction}
\label{sec:intro}
Novel view synthesis (NVS) of urban scenes is essential for autonomous driving applications~\cite{nerf,chen2025g3r,blocknerf}. Although NeRF-based approaches have shown promising results in urban environments~\cite{urbanradiancefield,f2nerf,unisim,miao2024edus}, their slow rendering speeds limit practical applications. Recently, newer methods have employed 3DGS for urban view synthesis~\cite{hugs,streetgaussian,khan2024autosplat,zhou2024drivinggaussian,chen2023periodic,hierarchicalgs}. However, they still require approximately one hour of training per urban scene.

To reduce training time while maintaining rendering efficiency, recent methods have explored feed-foward 3DGS predictions~\cite{charatan2024pixelsplat,chen2024mvsplat,szymanowicz2024splatter,smart2024splatt3r,szymanowicz2024flash3d,liu2025mvsgaussian,wewer2024latentsplat}.
These methods share a similar design, utilizing 2D CNN and transformer architectures to predict pixel-aligned 3D Gaussians from 2D reference images. Despite promising NVS results, they encounter challenges in street scenes: 1) These methods rely heavily on feature matching, constructing local frustums under each reference view to predict per-pixel Gaussians. However, driving datasets typically feature small parallax angles and texture-less regions, making depth prediction through feature matching difficult. 2) The pixel-aligned design can lead to duplicate and inconsistent Gaussians in overlapping image regions~(\eg the multi-layer surfaces), resulting in ghosting artifacts and increased memory usage, as illustrated in \cref{fig:teaser}.

In this paper, we propose Efficient Volume-based Gaussian Splatting (\method) for fast reconstruction and real-time rendering of street view images.  Unlike existing methods that predict per-pixel aligned Gaussians, our approach directly predicts 3D Gaussians over multiple frames, where the geometry attributes are learned from 3D context and the color is predicted based on 2D texture information. Specifically, given a sequence of sparse images, we initialize our model by accumulating their monocular depth predictions into a noisy point cloud. This point cloud is then processed by a generalizable 3D-CNN, which decodes geometry attributes to transform the noisy point cloud into 3D Gaussians by predicting position offsets, scale, rotation, and opacity. While the 3D-CNN enables accurate geometry prediction, it struggles to capture high-frequency appearance details due to its inherent smoothness bias.
To address this, we introduce an occlusion-aware, image-based rendering (IBR) module that predicts Gaussian colors from aggregated 2D features for regions within the volume’s range. For feed-forward inference over unbounded scenes, we further model distant views and the sky with 3D Gaussians positioned on a distant hemisphere.

Our contributions can be summarized as follows:
1) We introduce a novel feed-forward reconstruction method tailored to unbounded driving scenes, employing two distinct generalizable components for the foreground and background. It achieves efficient reconstruction and real-time rendering from sparse vehicle-mounted cameras.
2) We predict Gaussian geometric and appearance properties separately using a global volume-based representation and an occlusion-aware IBR module. These enhancements enable high-quality reconstruction of urban scenes, particularly in the presence of occlusions, while also reducing memory consumption.
Experimental results show that our method outperforms other generalizable baselines on the KITTI-360 dataset and exhibits promising zero-short generalization abilities on the Waymo dataset.

\begin{figure*}
  \centering
   \includegraphics[width=0.9\linewidth]{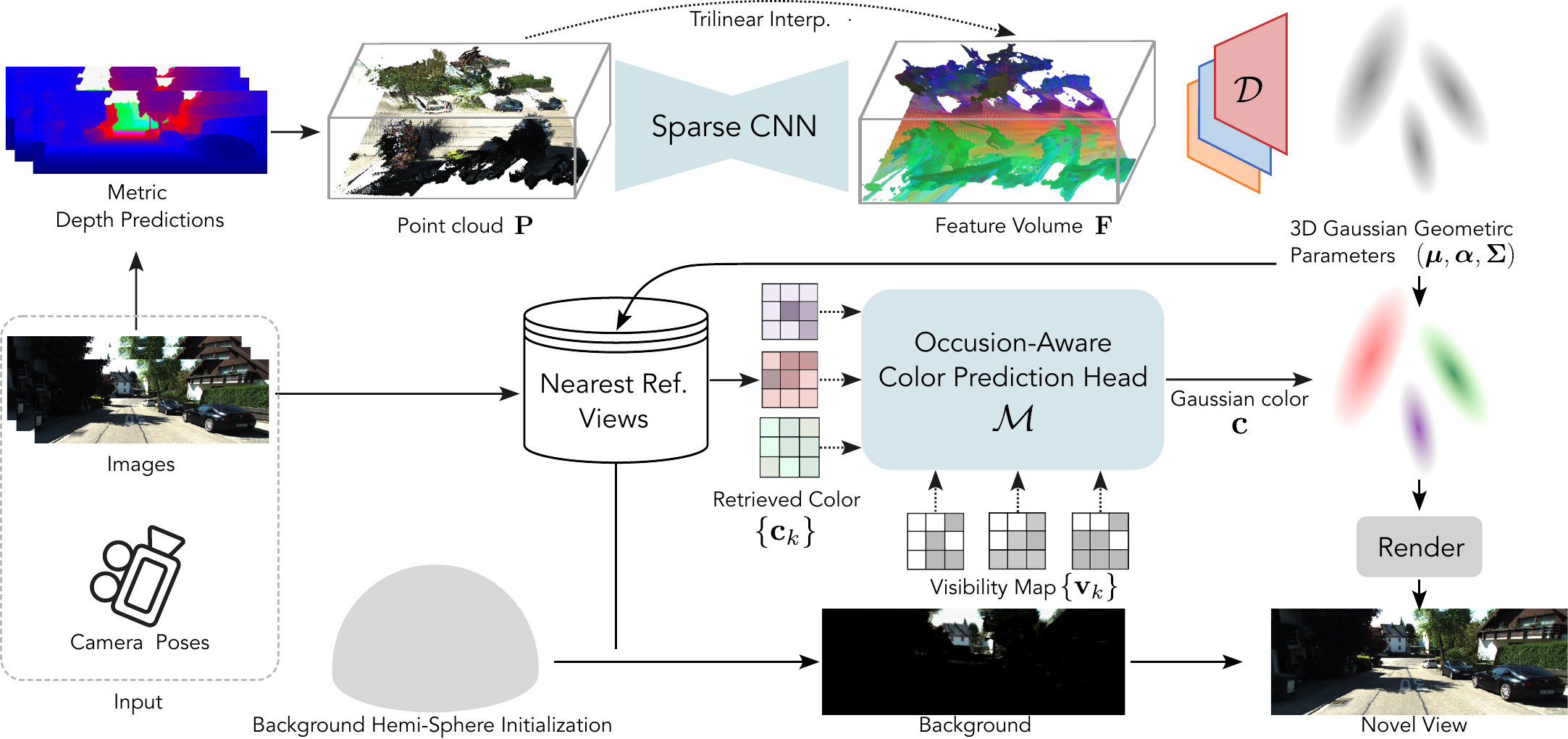}
   \caption{\textbf{Method}. EVolSplat learns to predict 3D Gaussians of urban scenes in a feed-forward manner. Given a set of posed images $\{I_n\}_{i=1}^N$, we first leverage off-the-shelf metric depth estimators to provide depth estimations $\{D_n\}_{n=1}^N$. The depth maps are unprojected and accumulated into a global point cloud $\bP$, which is fed into a sparse 3D CNN for extracting a feature volume $\bF$. We leverage the 3D context of $\bF$ to predict the geometry attributes of 3D Gaussians, including their center $\bmu$, opacity $\balpha$, and covariance $\bSigma$. Furthermore, we project the 3D Gaussians to the nearest reference views to retrieve 2D context, including color window $\{\bc_k\}_{k=1}^K$ and visibility maps $\{\bv_k\}_{k=1}^K$ to decode their color. To model far regions, we propose a generalizable hemisphere Gaussian model, where the geometry is fixed and the color is predicted in a similar manner as the foreground volume. }
   \label{fig: pipeline}
   \vspace{-0.5cm}
\end{figure*}

\section{Related Work}
\label{sec:related}
\boldparagraph{Novel View Synthesis for Driving Scenes}
The rapid development of radiance field techniques\cite{3dgs,nerf} have significantly advanced the development of novel view synthesis for driving scenes. These approach represent street scenes as implicit neural fields~\cite{streetsurf,snerf,urbanradiancefield,wu2023mars,unisim,cheng2023uc}  or anisotropic 3D Gaussians ~\cite{hugs,streetgaussian,khan2024autosplat,zhou2024drivinggaussian,chen2023periodic,hierarchicalgs}, achieving expressive synthesis quality.
A few works~\cite{snerf,urbanradiancefield,unisim} leverage sparse vehicle-mounted LiDAR sensors to boost the training and learn a robust geometry. Other works also incorporate additional semantic and geometry cues as priors or supervision to enhance scene comprehension, including semantic understanding~\cite{wu2023mars, fu2023panopticnerf,pnf}, geometric constraints~\cite{khan2024autosplat, streetsurf} and static-dynamic decomposition~\cite{emernerf, zhou2024drivinggaussian}. However, these methods require expensive per-scene optimization. Our generalizable method, in contrast, aims to perform efficient reconstruction on novel street scenes.

\boldparagraph{Feed-Forward Scene Reconstruction}
To perform generalizable reconstruction, researchers train neural networks across large-scale datasets to gain domain-specific prior knowledge. These methods~\cite{pixelnerf,charatan2024pixelsplat,chen2024mvsplat,zhang2024transplat,gslrm,szymanowicz2024splatter} have evolved to work with sparse image sets and can directly reconstruct scenes via fast feed-forward inference. Generalizable NeRF performs ray-based rendering and estimates the 3D radiance field of scenes, which enables superior cross-dataset generalization capabilities in both small~\cite{mvsnerf,ibrnet,pixelnerf} and large baseline~\cite{xu2023murf,du2023learning} settings. However, NeRF based methods are notoriously slow to render. Another line of work directly predicts pixel-aligned 3D Gaussians from sparse reference images~\cite{szymanowicz2024flash3d,szymanowicz2024splatter,gslrm,chen2024mvsplat,wewer2024latentsplat,chen2025g3r}. These methods either utilize an epipolar transformer~\cite{charatan2024pixelsplat,wewer2024latentsplat} or construct local cost volumes~\cite{chen2024mvsplat,zhang2024transplat,liu2025mvsgaussian} for learning geometry. However, most of these methods focus on small-scale scenes and require overlaps in the input images, making them difficult to apply in driving scenes with large camera movements and small parallax. Our method predicts the global scene representation instead of per-pixel Gaussian by accumulating and refining depth information in 3D space. This improves global consistency and reduces ghosting artifacts when rendering new images from new viewpoints. In related work, Flash3D~\cite{szymanowicz2024flash3d} and DrivingForward~\cite{tian2024drivingforward} also uses monocular depth predictions to initialize Gaussian locations but regresses pixel-aligned Gaussian parameters from a 2D feature map. In contrast, we downsample the initial primitives to reduce computation and learn Gaussian geometry in 3D, better incorporating neighbor information.

\boldparagraph{Point-based Novel View Synthesis}
Point clouds capture the geometric structure of a scene, making them a popular representation for enhancing NVS quality. 
While points can be augmented with learnable descriptors~\cite{sun2025pointnerf++,ruckert2022adop}, they often lead to rendering holes. To address this, some methods extend point primitives to continuous 3D Gaussians~\cite{3dgs}, or use neural networks for post-processing to fill gaps~\cite{ruckert2022adop,aliev2020neural,franke2023vet}. Although these approaches are efficient for rendering complex geometry, they rely on per-scene optimization. In contrast, we propose a generalizable model based on point cloud data.

A few works~\cite{huang2023ponder,yang2023unipad}
pretrain an encoder-decoder architecture with skip connections to extract the spatial and local features of point clouds for downstream tasks like object detection and semantic segmentation, but do not consider unbounded street views. PointNeRF~\cite{xu2022point} and EDUS~\cite{miao2024edus} learn implicit radiance fields from dense input point clouds, limiting their practical applicability due to slow rendering. Our method achieves real-time rendering by feed-forward prediction of Gaussian primitives.

\section{Method}
\label{sec:method}

Our goal is to learn a feed-forward 3DGS model for efficient street scene reconstruction from sparse input views, with optional fine-tuning for further improvement. An overview of~\method~is provided in \figref{fig: pipeline}. We decompose the entire street scene into foreground and background, and represent them via two independent GS prediction models $\mathcal{G}_{fg}$ and  $\mathcal{G}_{bg}$. The term \textbf{foreground} denotes the region within a predefined volume that encompasses the vehicle's trajectory, while \textbf{background} refers to the distant scenery and sky outside this volume. 

 We begin with a set of sparse images and exploit an off-the-shelf depth model to initialize global primitives (\secref{sec:pointcloud generation}), which are transformed into an encoding volume to predict the Gaussian geometry and appearance parameters (\secref{sec:encoding geometry volume}). We further train a generalizable hemisphere model to represent the background (\secref{sec:background}). Finally, the foreground and background are composed to reconstruct the full image (\secref{sec:details}).

\subsection{Preliminary}
\label{sec:Preliminary}
3D Gaussian Splatting~\cite{3dgs} explicitly parameterizes the 3D radiance field of the underlying scene as a collection of 3D Gaussians pritmitives $\mathcal{G}=\left\{\left(\boldsymbol{\mu}_i,\boldsymbol{\alpha}_i,\boldsymbol{\Sigma}_i, \boldsymbol{c}_i\right)\right\}_{i=1}^G$, with attributes: a mean position $\boldsymbol{\mu_i}$, an opacity $\boldsymbol{\alpha_i}$, a covariance matrix $\boldsymbol{\Sigma_i}$ and view-dependent color $\boldsymbol{c_i}$ (computed by spherical harmonics). This efficient representation avoids expensive volumetric sampling and enables high-speed rendering. To render the image from a particular viewpoint, 3DGS employs the tile-based rasterizer for Gaussian splats to pre-sort and blend the ordered primitives using differentiable volumetric rendering:
\begin{equation}
\boldsymbol{c}=\sum_{i \in G}  \boldsymbol{c}_i \boldsymbol{\alpha}_i \prod_{j=1}^{i-1}\left(1-\boldsymbol{\alpha}_j\right)
\end{equation}

\subsection{Global Point Cloud Initialization}
\label{sec:pointcloud generation}
\boldparagraph{Depth Estimation}
We integrate depth priors into the vanilla 3DGS framework to model the foreground. Given $N$ input images $\{I_{n}\}_{n=1}^{N}$ for a single scene, we leverage a pre-trained depth model to predict metric depth maps $\{D_{n}\}_{n=1}^{N}$ for each frame (\eg Metric3D~\cite{yin2023metric3d}, Unidepth~\cite{piccinelli2024unidepth}). Inspired by~\cite{cheng2023uc,miao2024edus}, we also apply a depth consistency check to remove inconsistent depth values, and use 3D filters to remove clear floating artifacts in the unproject point clouds. The obtained metric depth information enables us to unproject per-frame monocular depth predictions to global world coordinates. We provide the depth consistency check details in the Supplementary material.

\boldparagraph{Global Point Cloud Construction}
We accumulate the predicted metric depth maps $\{D_{n}\}_{n=1}^{N}$ into a global point cloud. Specifically, we lift the RGB images $\{I_{n}\}_{n=1}^{N}$ into 3D space and accumulate them to construct a global scene point cloud $\mathcal{P} \in \mathbb{R}^{N_p \times 3}$ in world coordinates with the calibrated intrinsic matrices $\left\{K_n\right\}_{n=1}^N$ and extrinsic matrices$\left\{T_n = R_n \mid t_n\right\}_{n=1}^N$:
\begin{equation}
\mathcal{P}=\bigcup^N \mathbf{\pi}^{-1}\left(I_n, D_n, \mathbf{T}_n, \mathbf{K}_n\right)
\end{equation}
where $\pi^{-1}$ denotes pixel unprojection. Previous pixel-aligned 3DGS methods will inevitably generate redundant and inconsistent primitives in overlapping region. To address this problem, we apply a uniform filter for the raw points $\mathcal{P}$ in 3D space to delete duplicate primitives while keeping the scene structure. Note that the processed primitives still contain noise, so we apply a statistical filter to remove floaters and explain how to further refine their locations in the next section.

\subsection{Volume-Based Gaussian Generation}
\label{sec:encoding geometry volume}
\boldparagraph{Construct Neural Volume}
We draw inspiration from~\cite{neo360,huang2023ponder,chen2024lara} and train a generalizable 3D latent neural volume for modeling prior knowledge across scenes. More specifically, We construct the sparse tensor from initialized noisy primitives $\mathcal{P}$ and employ an efficient sparse U-Net style 3DCNN $\psi^\text{3D}$ to aggregate the neural feature volume $\bF \in \mathbb{R}^{H \times W \times D \times C}$, where $H \times W \times D $ denotes the spatial resolution and $C$ denotes the feature dimension. The sparse 3D ConvNet follows an encoder-decoder structure with a bottleneck of low spatial dimension. We additionally add skip connections to keep both local and global information.
\begin{equation}
\bF=\psi^\text{3D}(\mathcal{P}).
\end{equation}
Different from other feed-forward GS works using a 2D image-to-image neural network to encode the scene, this 3D latent representation helps integrate local and global information in different network layers for faithful reconstructions. Note that the output feature volume $\bF$ is aligned with the world coordinate system.

\boldparagraph{Gaussian Geometry Parameters Prediction}
We decode the geometry attributes of $N_p$ 3D Gaussians based on the feature volume $\bF$, taking the scene point cloud $\cP$ as the initial Gaussian centers, i.e., $\mu_i^{init}=\bp_i \in \cP$. 
For each $\mu_i^{init}$, we query a latent feature from $\bF$ using trilinear interpolation, and map it to a position offset $\Delta \mu_i$, opacity $\balpha_i$, and covariance matrix $\bSigma_i$ based on three independent MLP heads, respectively.

Considering that the network is supposed to gradually move the 3D Gaussians to the correct locations over the training process, we aim to use the updated locations $\mu_i + \Delta \mu_i$ to query the feature volume to decode $\balpha_i$ and $\bSigma_i$. This encourages the network to predict $\balpha_i$ and $\bSigma_i$ at a more accurate position, i.e., near the object surface. However, we don't have a good estimation of the $\Delta \mu_i$ before training. Therefore, we build an updating rule, with which $\Delta \mu_i$ converges at the end of the training. 
Specifically, we keep track of the network's prediction of $\Delta \mu_i$ from the last iteration as $\Delta \mu_i^{prev}$. For each iteration, we query $\bF$ at $\mu_i^{init} + \Delta_i^{prev}$, obtaining a feature vector $\bff_i$:
\begin{align}
   \bff_i = & \bF (\mu_i^{init} + \Delta \mu_i^{prev}) \label{eq:offset1} \\
   \balpha_i = & \mathcal{D}_{opa}(\bff_i),~~~\bSigma_i=\mathcal{D}_{cov}(\bff_i) \\
   \Delta \mu_i = & \textit{Tanh}(\mathcal{D}_{pos}(\bff_i))\cdot v_\text{size}, ~~~\mu_i= \mu_{init} + \Delta \mu_{i}  \label{eq:offset2}
\end{align}
where $\mathcal{D}_{opa}$, $\mathcal{D}_{cov}$ and $\mathcal{D}_{pos}$ denote MLP heads, and $\mu_i, \balpha_i$ and $\bSigma_i$ are subsequently used for rendering. Here, the output of $\mathcal{D}_{pos}$ is passed through a $\textit{Tanh()}$ activation function, and then scaled by the voxel size $v_\text{size}$ to yield the final adjustment. 
Note that this leads to a recursive formulation of $\Delta \mu_i$ where its value depends on the output of the previous estimations. We show that $\Delta \mu_i$ is stabilized to a stationary point at infinite training iterations in the supplementary. During inference, we initialize $\Delta \mu_i^{prev}=0$, recursively execute \cref{eq:offset1} and \cref{eq:offset2}, and use the converged point to query $\balpha_i$ and $\bSigma_i$. We empirically find that it is sufficient to only execute the recursion twice to obtain a converged $\Delta \mu_i$.

As shown in our ablation experiment, the inclusion of offset allows our method to compensate for noisy positions and enhances the representational capacity.

\begin{figure}[tb]
  \centering
   \includegraphics[width=0.8\linewidth]{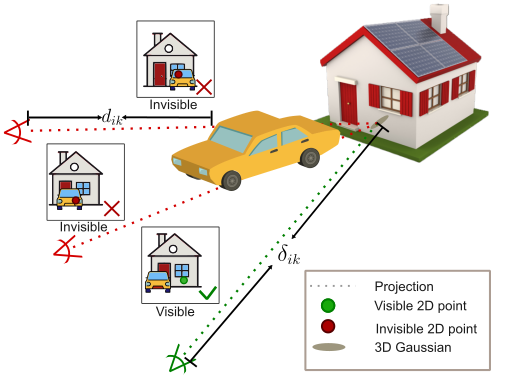}
   \vspace{-0.4cm}
   \caption{\textbf{Occlusion Illustration}. One Gaussian in 3D space may retrieve inaccurate color information from 2D reference images due to occlusions. ~\method~comprises geometric priors to reduce the impact of invisible colors to enhance rendering quality.   }
   \label{fig:occlusion}
   \vspace{-0.4cm}
\end{figure}

\boldparagraph{Occlusion-Aware IBR-based Color Prediction}
Our foreground volume-based representation downsamples the initial primitives, which risks losing high-frequency appearance details. To trade off memory consumption and model high-detail appearance, we introduce the image-based rendering technique to learn Gaussian appearance. However, two challenges arise when applying IBR to urban scenes: 1) coarse Gaussian centers initialization results in inaccurate 2D projection, and 2) the retrieved color is inconsistent for a Gaussian due to occlusions, as illustrated in \figref{fig:occlusion}. To address these issues, we extend the projection location into a local sample window and leverage the depth priors to eliminate the invisibility projection.

Specially, for one Gaussian primitive $\mathcal{G}_i$, we project its center $\mathbf{\mu}_i$ to the nearby $K$ reference frames and sample the local color $\left\{\bc_{ik}  \in \nR^{W\times W \times 3}\right\}_{k=1}^K$ with a predefined $W \times W$ window centered at the 2D projection of the 3D Gaussian. This strategy allows for integrating additional 2D texture information, enhancing the model's robustness against noisy 3D Gaussian locations.  We experimentally observe that $W=3$ is an effective choice, as demonstrated in the ablation study. 

To perform an occlusion check, we bilinearly retrieve the monocular depth $D_r$ as pseudo ground truth and estimate the visibility maps $\{\bv_{i k} \in \nR^{W\times W}\}$ based on the difference between the projected depth and $D_r$. Specifically, let $\delta_{i k}$ denote the distance between one Gaussian center $\mathbf{\mu}_i$ and the projected camera center $\mathbf{t}_k$, i.e., $\delta_{i k}=\left\|\mathbf{\mu}_i-\mathbf{t}_k\right\|_2$. We estimate the visibility based on the depth window $\bd_{i k} \in \nR^{W\times W}$ corresponding to $\bc_{i k}$, where $\bv_{i k} =(\delta_{i k}-\bd_{i k})/\delta_{i k}$.
With the powerful input depth priors as guidance, our method effectively distinguishes the occlusion-caused feature, allowing it to focus on the visible views. Instead of removing invisible projections that require a manually determined threshold, we propose an aggregation module $\mathcal{M}$ to derive the Gaussian color $\bc_i$ by combining the local window colors with their corresponding visibility terms. We implement $\mathcal{M}$ as a three-layer MLP of width 64 and adopt the spherical harmonics coefficients (SH) to represent the color.  Inspired by~\cite{lu2024scaffold}, we further take as input the distance and relative direction between Gaussian and target camera to make our model structure-aware and facilitate better image synthesis quality.
\begin{equation}
\bc_i=\mathcal{M}\left(\{\bc_{i k}, \bv_{i k},\delta_{i k},\overrightarrow{\mathbf{\theta}}_{i k}\}_{k=1}^{K}\right), 
\overrightarrow{\mathbf{\theta}}_{i k}=\frac{\mathbf{\mu}_i-\mathbf{t}_k}{\left\|\mathbf{\mu}_i-\mathbf{t}_k\right\|_2}
\end{equation}

\subsection{ Generalizable Background Model}
\label{sec:background}
In this section, we introduce our generalizable background design tailored for driving scenes to model infinite sky and distant landscapes (often 100 meters away). Obviously, the volume-based representation covers a limited street scene and is insufficient for the unbounded region. Previous work~\cite{chen2023periodic,ye2024gaustudio,xu2024splatfacto,urbanradiancefield} construct environment map conditioned on input direction for a special scene, hence struggle to handle multiple scenes simultaneously. 

Given that the background region is far away,
we propose a generalizable hemisphere Gaussian model to represent the background. We uniform sampling points over the hemisphere and project all these points to reference images to query the 2D color $\{\bc_{k}\}_{k=1}^K$ on the reference image. This background hemisphere has a fixed radius $r_{bg}$ that moves along with the camera motion such that the relative distance is fixed \wrt the target view. To convert each point on this hemisphere into Gaussian primitives $\mathcal{G}_{bg}$, we employ a lightweight two-layer MLP $\mathcal{M}_{bg}$ to learn its color and scale from $\{\bc_{k}\}_{k=1}^K$. 
\begin{equation}
\bc_{bg},\bs_{bg} = \mathcal{M}_{bg}(\{\bc_{k}\}_{k=1}^K)
\end{equation}
Note that we use fixed parameters for its rotation and opacity, see supplementary for more details. This simple strategy allows our model to reconstruct the full image via efficient 3DGS rasterization while keeping the desirable background rendering.

\subsection{Training \& Fine-tuning}
\label{sec:details}

\boldparagraph{Training Loss} We jointly train our scene representation, foreground volume, and hemisphere background models. We apply L1 and SSIM losses between rendered and observed images for supervision of RGB rendering.
\begin{equation}
\mathcal{L}_{\mathrm{rgb}}=\left(1-\lambda_r\right) \mathcal{L}_1+\lambda_r \mathcal{L}_{\mathrm{ssim}}
\end{equation}
Following\cite{streetsurf,streetgaussian}, we additionally adopt entropy regularization loss on the foreground accumulated alpha value $\mathbf{O}_{\text {fg}}$ to encourage opaque rendering.
\begin{equation}
\mathcal{L}_{\text {entropy}}=-\sum\left(\mathbf{O}_{\text {fg}}\log \mathbf{O}_{\text {fg}}+\left(1-\mathbf{O}_{\text {fg}}\right) \log \left(1-\mathbf{O}_{\text {fg}}\right)\right)
\end{equation}

The total loss can be summarized as follows:
\begin{equation}
\mathcal{L}=\left(1-\lambda_r\right) \mathcal{L}_1+\lambda_r \mathcal{L}_{\mathrm{ssim}}+\lambda_e \mathcal{L}_{\text {entropy}}
\end{equation}

\boldparagraph{Fine-tuning}
Once fine-tuning is applied, our~\method can achieve photorealism on par with or surpassing other methods, leveraging powerful pretrained weights. Specifically, we first predicts a set of 3D Gaussian primitives for initialization via a direct feed-forward process, where both the geometry and color attributes are generated. 
The fine-tuning process follows the vanilla 3DGS~\cite{3dgs}, where we optimize all geometry and color attributes directly. We also enable the growing and pruning of 3D primitives during fine-tuning.
During fine-tuning, the number of Gaussians is significantly reduced, as the initial feed-forward Gaussians are redundant. As a result, our model maintains high fidelity while reducing memory consumption. We demonstrate that the pretrained priors accelerate network training and enable faster convergence compared to other test-time optimization methods.

\boldparagraph{Implementation Details}
We train~\method~on multiple scenes using Adam optimizers~\cite{kingma2014adam} with learning rate $1 \times 10^{-3}$. 
We use torchsparse~\cite{tang2022torchsparse} as the implementation of 3D sparse convolution and choose gsplat~\cite{ye2024gsplatopensourcelibrarygaussian} as our Gaussian rasterization library. We set the SH degree to 1 for simplicity, following StreetGaussian~\cite{streetgaussian}. During training, we set $\lambda_r = 0.2$, $\lambda_e = 0.1$ as loss weights in our work. For each iteration, our model randomly selects a single image from a random scene as supervision.

\begin{table*}[ht]
\centering
\small
\setlength{\tabcolsep}{11pt}
\begin{tabular}{@{}l|ccc|ccc|cc@{}}
\toprule
\multirow{2}{*}{Method} & \multicolumn{3}{c|}{KITTI-$360$} & \multicolumn{3}{c|}{Waymo } & \multirow{2}{*}{FPS$\uparrow$} & \multirow{2}{*}{Mem.(GB)$\downarrow$}\\
             & PSNR(dB)$\uparrow$  & SSIM$\uparrow$ & LPIPS$\downarrow$ & PSNR(dB)$\downarrow$ & SSIM$\uparrow$    &  LPIPS$\downarrow$    \\ \midrule
MVSNeRF~\cite{mvsnerf}     &18.44  &0.638  &0.317  &17.86  &0.595  &0.433 &0.025  &12.03  \\
MuRF~\cite{xu2023murf}     &22.77 &0.780  &0.229 &23.33 &0.770 &0.269 &0.31 &26.45 \\
EDUS~\cite{miao2024edus}    & 22.13  & 0.745  & \textbf{0.178} &23.18  &0.745  &\textbf{0.164} & 0.14 & \textbf{5.75}    \\ 
\midrule
MVSplat~\cite{chen2024mvsplat}    &21.22  &0.695 &0.268 & 21.33  & 0.665  &0.308 &81.58 &16.14 \\
PixelSplat~\cite{charatan2024pixelsplat}   &19.41   &0.584  &0.357  & 16.65 &0.541  &0.579 &70.46  &26.75 \\

Ours          & \textbf{23.26} & \textbf{0.797} &0.179  &\textbf{23.43}  &\textbf{0.786}  &0.202 &\textbf{83.81} &10.41\\
\bottomrule
\end{tabular}
\vspace{-0.2cm}
\caption{\textbf{Quantitative results on KITTI-360 and Waymo datasets} with other generalizable methods. All models are trained on the KITTI-360 dataset using drop50\% sparsity level. Metrics are averaged on five validation scenes without any finetuning. Our~\method~generalizes better than all the baselines in terms of PSNR and SSIM on both KITTI-360 and Waymo open datasets. It is also worth noting that our method is more memory efficient compared with other 3DGS-based methods.}
\label{tab:feed forward}
\end{table*}

\section{Experiment}
\label{sec:experiment}
\subsection{Experimental Setup}
\vspace{-0.2cm}
\boldparagraph{Dataset and metrics}
We pretrain our model on the KITTI-360 dataset~\cite{liao2022kitti}.  In the case of training scenes, we collect 160 sequences from different geographic variations, with each sequence comprising 30 stereo images. We use five public validation sets from the KITTI-360 dataset, which has no overlapping with the training set.  Additionally, we access the zero-shot generalization performance on the public Waymo Open Dataset~\cite{waymo}. In our experiments, we apply 50\% drop rate for all training and validation scenarios to amplify the movement between adjacent cameras.

For novel view synthesis, we adopt existing evaluation protocols, including PSNR, SSIM, and LPIPS, for quantitative assessments.  Regarding the render efficiency, we report frames per second (FPS) and inference memory usage on the same device to ensure a fair comparison. 

\boldparagraph{Volume Setup}
In this paper, we construct the axis-aligned bounding box to partition the foreground.  We set the height (Y-axis) and width (X-axis) of the foreground volume to encompass objects in the field, with a height of $12.8\mathrm{m}$ and a width of $32\mathrm{m}$. The Z-axis is set to cover the vehicle’s forward trajectory with a length of $64\mathrm{m}$. The input point cloud is voxelized with a voxel size of $(0.1\mathrm{m}, 0.1\mathrm{m}, 0.1\mathrm{m})$, resulting in the volume dimension of $128 \times 320 \times 640$. Notably, our 3D CNN effectively handles \textit{arbitrary} volume size well during inference. Compared to KITTI-360, the Waymo dataset captures urban data with a wider field of view and clearer distant scenes. Therefore, we adjust the foreground volume range to $(50\mathrm{m}, 20\mathrm{m}, 128\mathrm{m})$ in our zero-shot inference experiment.

\boldparagraph{Baselines}
We compare~\method~with several representative feed-forward methods, including MVSNeRF~\cite{mvsnerf}, MuRF~\cite{xu2023murf}, EDUS~\cite{miao2024edus}, MVSplat~\cite{chen2024mvsplat} and PixelSplat\cite{charatan2024pixelsplat}. Unless stated otherwise, we train and evaluate all generalizable models using the same data as \method. Note that We train MVSplat and PixelSplat on NVIDIA V100 since their transformer-based architecture requires a large amount of GPU memory, making it hard to train on consumer GPU (e.g., RTX4090 which we use for \method) at a full resolution.  
Additionally, we compare the NVS quality and reconstruction time with other fast per-scene optimization methods: Nerfacto~\cite{tancik2023nerfstudio} and  3DGS~\cite{3dgs}.

\subsection{Comparison with Feed-Forward Methods}
\begin{figure*}[t!]
    \centering
    \small 
    \setlength{\tabcolsep}{0pt}
    \def\mywidth{5.6cm}
    \begin{tabular}{p{0.5cm}p{\mywidth}p{\mywidth}p{\mywidth}}

    \rotatebox{90}{MVSplat} & 
    \includegraphics[width=\mywidth]{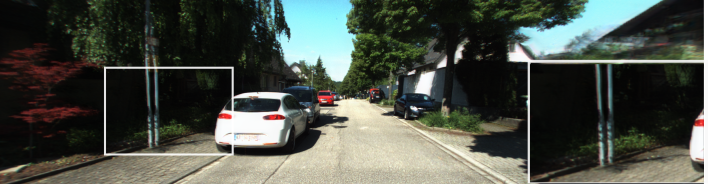}&
    \includegraphics[width=\mywidth]{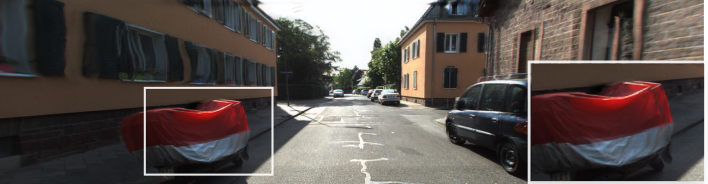}&
    \includegraphics[width=\mywidth]{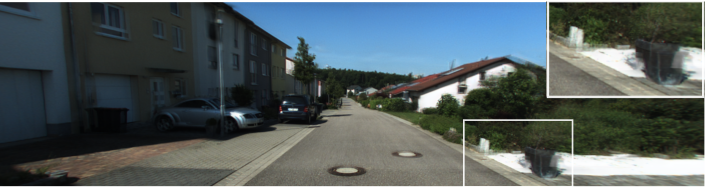} 
    \\
    \rotatebox{90}{EDUS} & 
    \includegraphics[width=\mywidth]{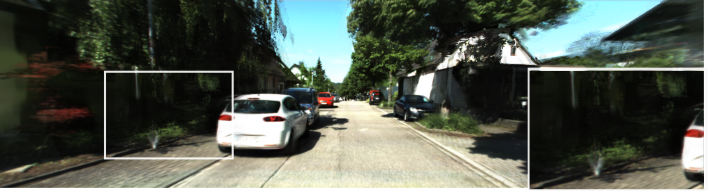}&
    \includegraphics[width=\mywidth]{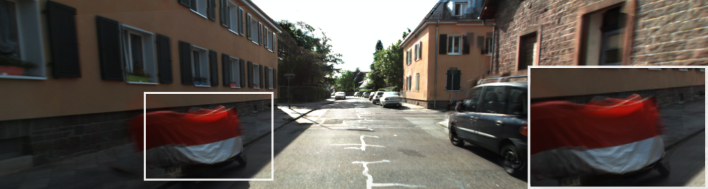}&
    \includegraphics[width=\mywidth]{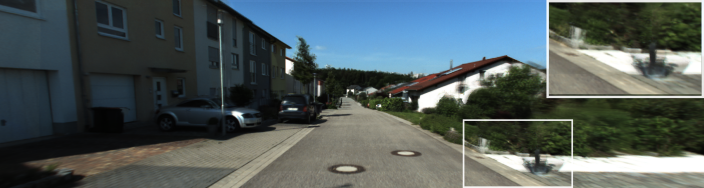} 
    \\
    \rotatebox{90}{Ours} & 
    \includegraphics[width=\mywidth]{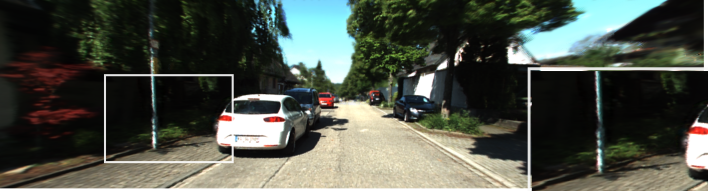}&
    \includegraphics[width=\mywidth]{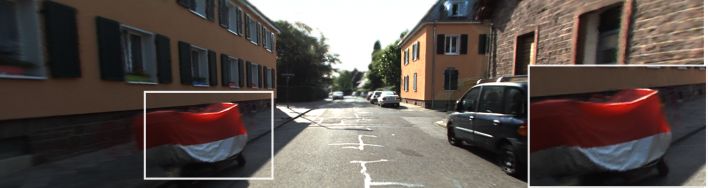}&
    \includegraphics[width=\mywidth]{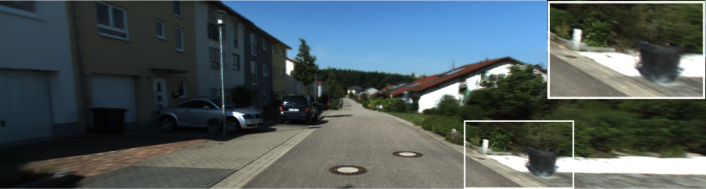} 
    \\
    \rotatebox{90}{~~~~~GT} & 
    \includegraphics[width=\mywidth]{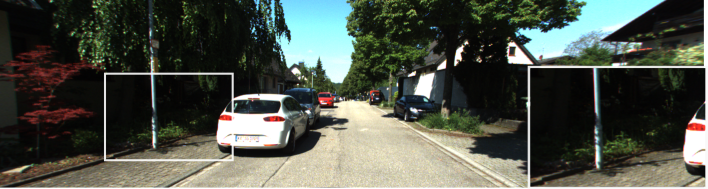}&
    \includegraphics[width=\mywidth]{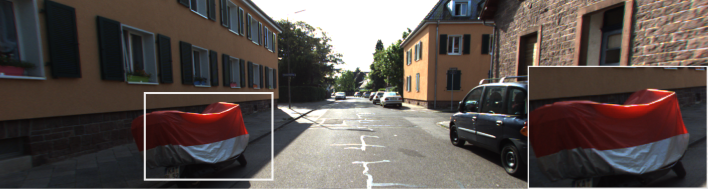}&
    \includegraphics[width=\mywidth]{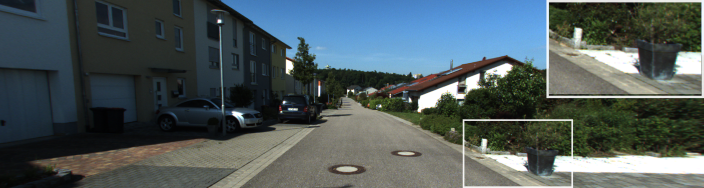} \\
    \end{tabular}
    \vspace{-0.2cm}
    \caption{{\bf Qualitative Comparison} with generalizable baselines on the KITTI-360 dataset.}  
    \label{fig:feed forward}
\end{figure*}

\vspace{-0.2cm}
\boldparagraph{Rendering Quality}
\cref{tab:feed forward} and \cref{fig:feed forward} present quantitative and qualitative comparison results for feed-forward inference, respectively. We adopt the $50\%$ drop rate following existing evaluation protocols \cite{liao2022kitti,wu2023mars, hugs}. Our proposed~\method~ achieves on par or better photorealism with the recent state-of-the-art NeRF-based method, while significantly improving rendering speed. Compared with feed-forward 3DGS prediction methods, our method achieves higher rendering quality. Specifically, MVSNeRF~\cite{mvsnerf} mainly focuses on object-level scenes and struggles with unbounded street views, leading to blurry images. We note that EDUS achieves good LPIPS, but has visual artifacts in thin structure and lower PSNR. MVSplat~\cite{chen2024mvsplat} estimates depth for each reference frame using a Transformer, which results in inconsistent 3D Gaussian primitives and causes ghost artifacts under large camera movements. 

\boldparagraph{Model Efficiency} 
We report the memory consumption during inference on a full-resolution $376\times1408$ image of KITTI-360 in \cref{tab:feed forward}. MuRF~\cite{xu2023murf} and PixelSplat~\cite{charatan2024pixelsplat} fail to train with full-resolutionon a consumer GPU (24GB) since their transformer architecture requires computation resources. In contrast, our method leverages an efficient sparse 3DCNN for generalizable reconstruction with only 10.41GB usage, demonstrating superior memory efficiency and practical utility.  Additionally, we note that EDUS~\cite{miao2024edus} consumes 5.75GB of memory for inference as its NeRF-based representation only casts 4096 rays each iteration,  resulting in lower memory usage but significantly slow rendering speed (0.14fps).

\boldparagraph{Zero-shot Inference on Waymo} 
\begin{figure*}[htbp]
    \centering
    \setlength{\tabcolsep}{0pt}
    \begin{tabular}{cc}
        \begin{minipage}{0.5\textwidth}
            \includegraphics[width=0.48\textwidth]{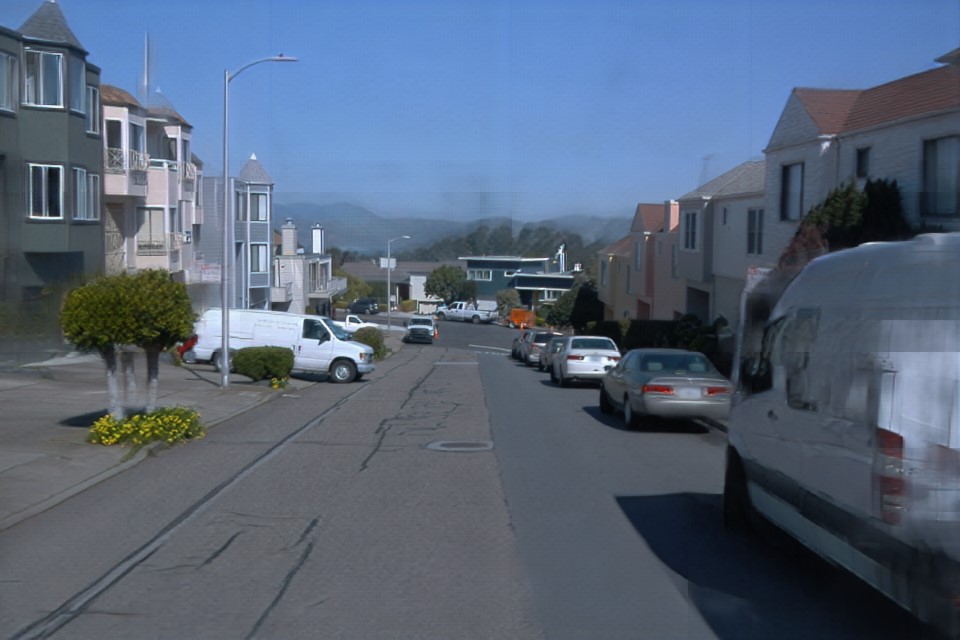} %
            \includegraphics[width=0.48\textwidth]{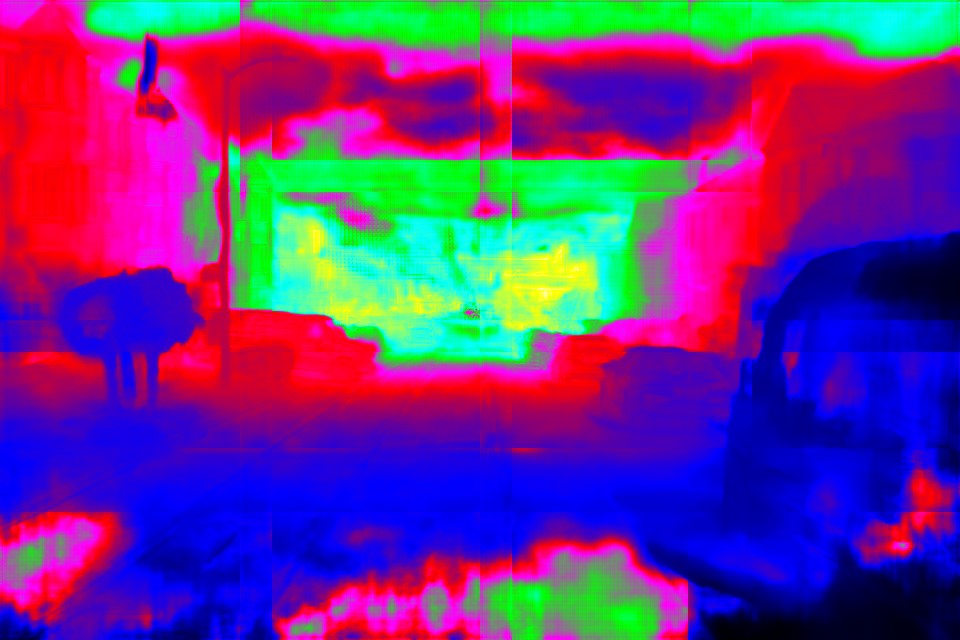} %
            \vspace{-0.25cm}
            \caption*{a) MuRF}
        \end{minipage} & 
        \begin{minipage}{0.5\textwidth}
            \includegraphics[width=0.48\textwidth]{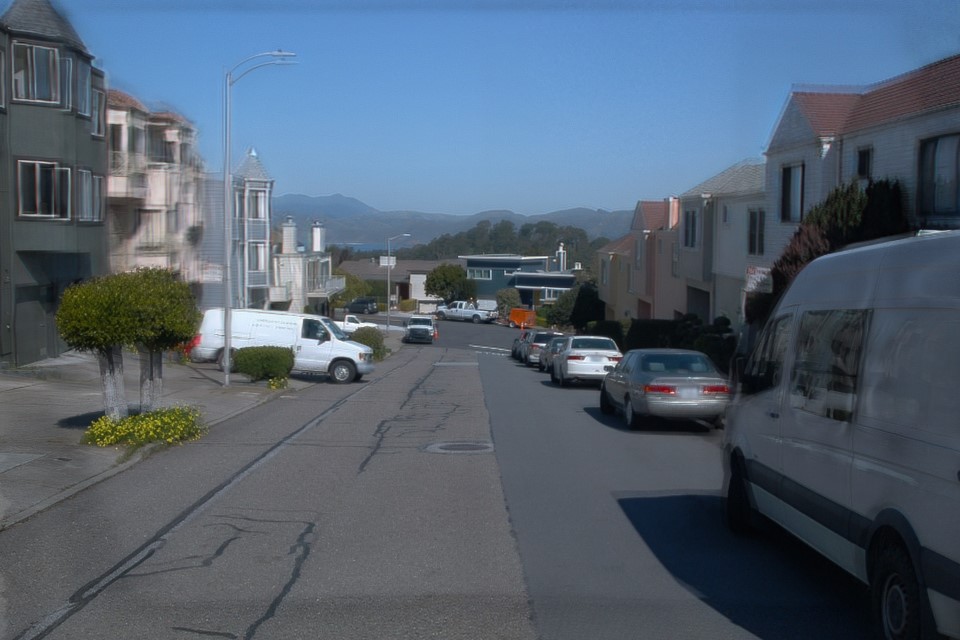} %
            \includegraphics[width=0.48\textwidth]{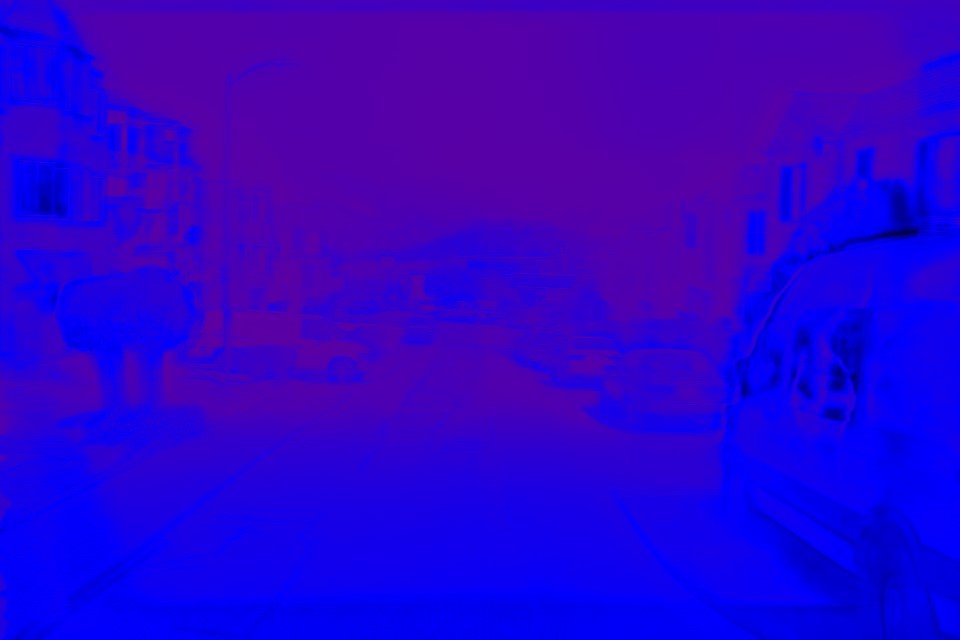} %
            \vspace{-0.25cm}
            \caption*{b) MVSplat}
        \end{minipage} 
        \vspace{0.1cm}  \\
        
        \begin{minipage}{0.5\textwidth}
            \includegraphics[width=0.48\textwidth]{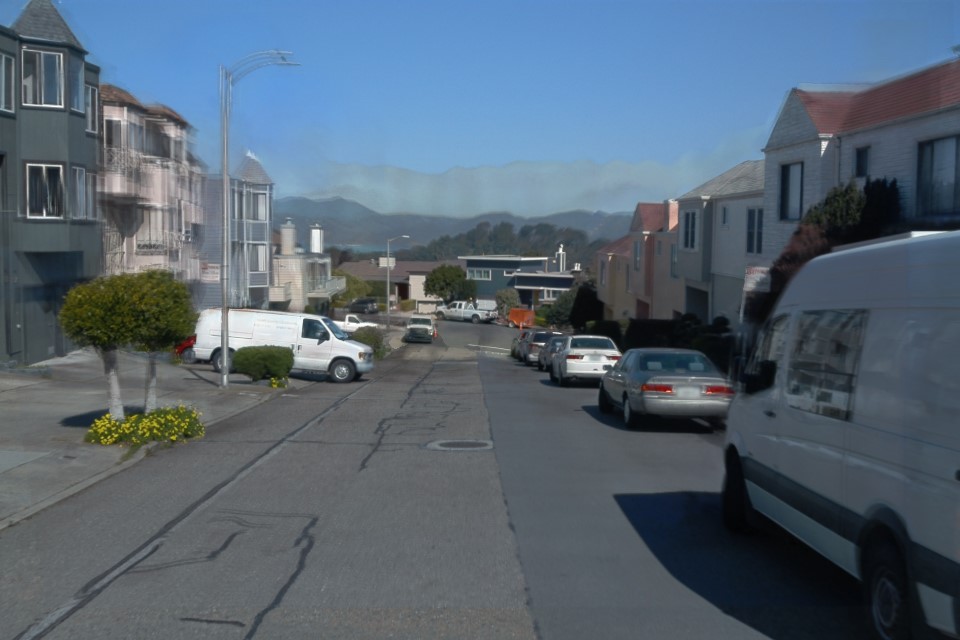} %
            \includegraphics[width=0.48\textwidth]{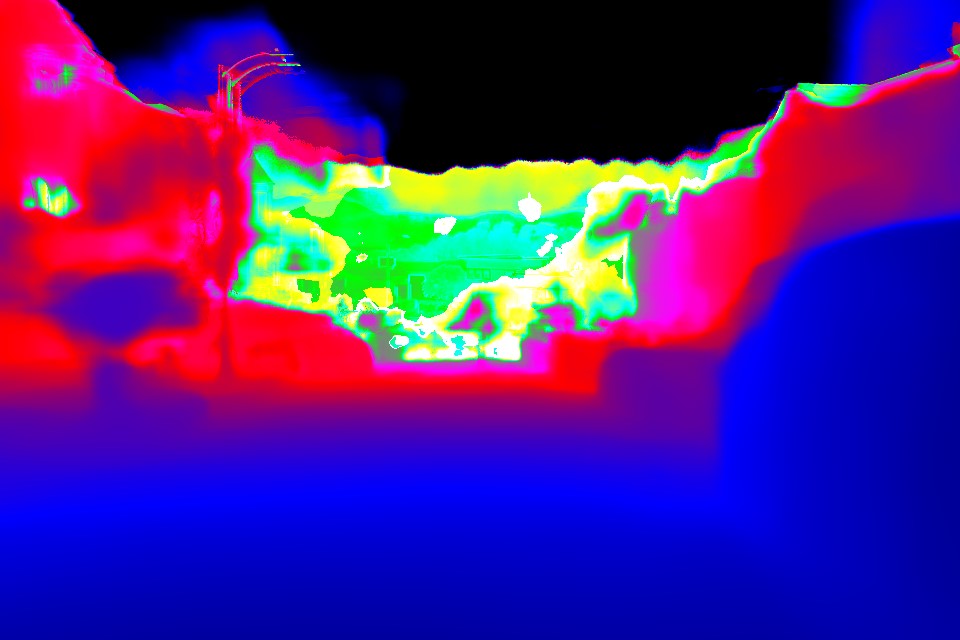} %
            \vspace{-0.25cm}
            \caption*{c) EDUS}
        \end{minipage} & 
        \begin{minipage}{0.5\textwidth}
            \includegraphics[width=0.48\textwidth]{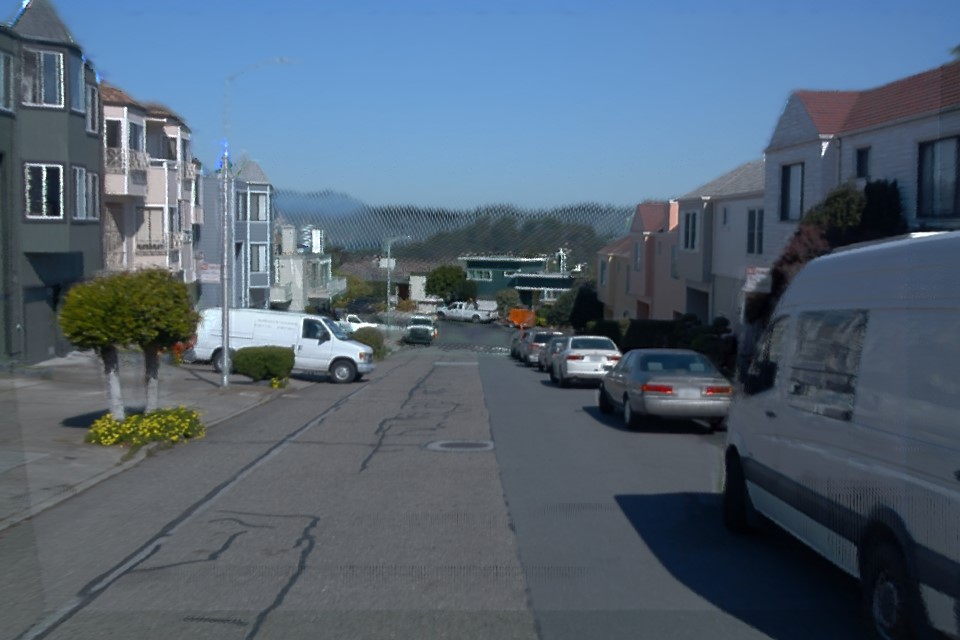} %
            \includegraphics[width=0.48\textwidth]{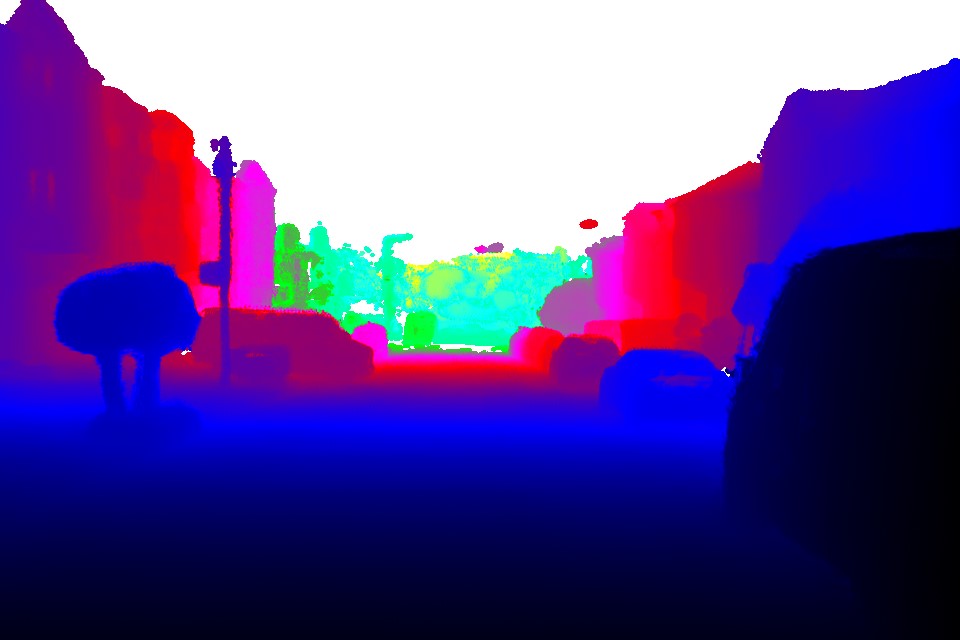} %
            \vspace{-0.25cm}
            \caption*{d) Ours}
        \end{minipage} 
        \vspace{0.1cm}  \\
    \end{tabular}
    \vspace{-0.2cm}
    \caption{\textbf {Qualitative Comparisons} with baselines for zero-shot inference on the Waymo Open dataset.}
    \label{fig:waymoQualitive}
\end{figure*}

To further verify the generalization capability, we evaluate different methods on an unseen dataset, the Waymo Open dataset~\cite{waymo}.
This is an out-of-domain setting for all methods as the models are trained on the KITTI-360 dataset. We set the image resolution to $640 \times 960 $ and evaluate the rendering quality and predicted geometry, see \cref{tab:feed forward} and \cref{fig:waymoQualitive}. As can be seen, our~\method~ achieves state-of-the-art performance  (PSNR: 23.43dB) on the unseen dataset. We observe that MuRF~\cite{xu2023murf} also achieves promising rendering results on Waymo, but it struggles with poor geometric reconstruction in texture-less areas, such as road, due to its strong reliance on feature mapping (as shown in the \cref{fig:waymoQualitive}), which limits its practical applicability on autonomous driving scenes. In contrast, our method benefits from the powerful geometric priors, demonstrating superior zero-shot generalization performance.

\subsection{Comparison with Optimization-based Methods}
\begin{figure}[tb]
  \centering
   \includegraphics[width=\linewidth]{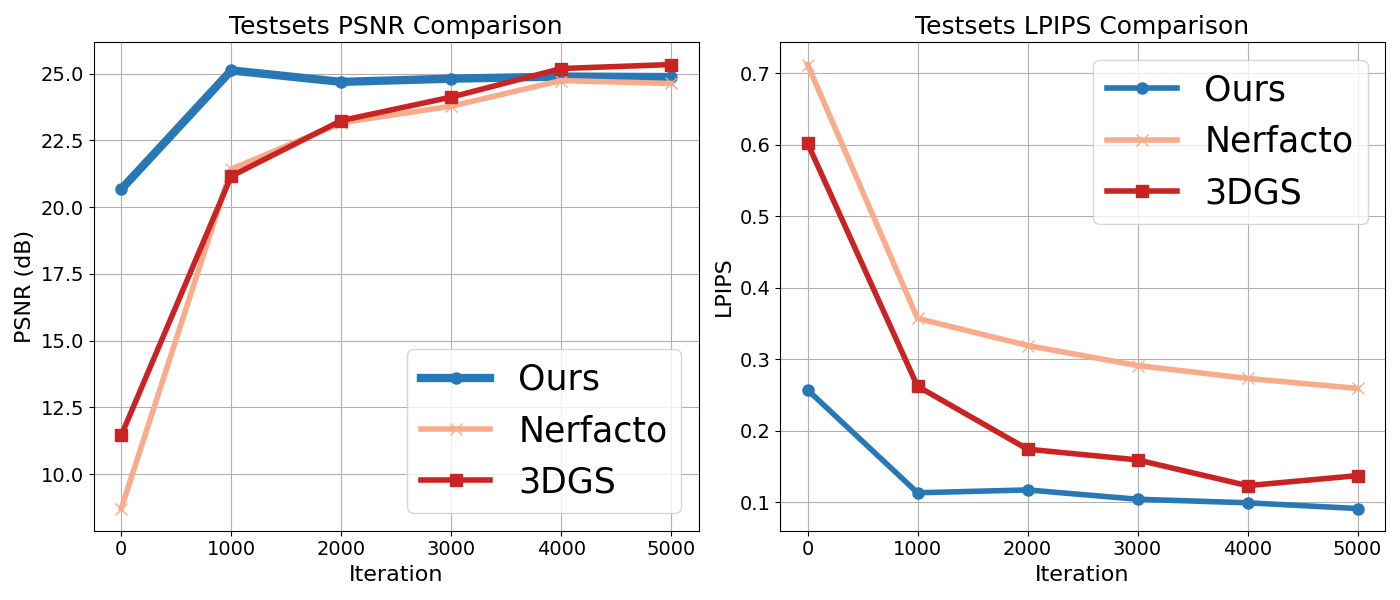}
   \vspace{-0.7cm}
   \caption{\textbf{Comparison with Optimization-based Methods.} We show PSNR and LPIPS on the test set at different training steps. Compared with test-time optimization baselines, our method with generalizable priors converges faster and achieves better LPIPS.}
   \label{fig:fineuning}
   \vspace{-0.4cm}
\end{figure}

We further compare the finetuning results with other test-time optimization methods and report the PSNR and LPIPS curve on the same test set in \cref{fig:fineuning}. 3DGS~\cite{3dgs} and Nerfacto~\cite{tancik2023nerfstudio} reconstruct the scenes from scratch, requiring longer training time. By leveraging pretrained weights, our model converges with fewer training steps (steps $<$ 1000).
Note that our method outperforms 3DGS in terms of LPIPS and has comparable PSNR after convergence.
\begin{table}[t]
\centering
\small
\setlength{\tabcolsep}{10pt}
\begin{tabular}{@{}lccc@{}} 
\toprule
& PSNR$\uparrow$  & SSIM$\uparrow$  & LPIPS$\downarrow$ \\
\cline{2-4}
w/o IBR &21.06 &0.706 &0.274  \\
w/o refine offset &22.76 &0.780  &0.195  \\
w/o occlusion check $ \bv_k$  &22.54  &0.781 &0.197   \\
Full model &22.83  &0.786  &0.190  \\
\bottomrule
\end{tabular}
\vspace{-0.2cm}
\caption{\textbf{Ablation Analysis} on KITTI-360 dataset.}
\vspace{-0.2cm}
\label{tab:ablation}
\end{table}

\subsection{Ablation study and analysis}
\begin{figure}[t]
     \centering
     \begin{subfigure}[h]{0.45\linewidth}
         \centering
         \includegraphics[width=\textwidth]{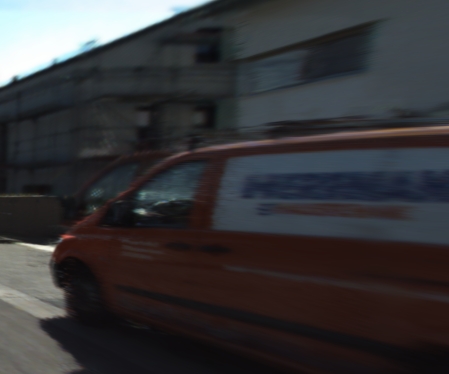}
        \caption{w/o IBR}
        \label{fig:wo ibr}
     \end{subfigure}
     \begin{subfigure}[h]{0.45\linewidth}
         \centering
         \includegraphics[width=\textwidth]{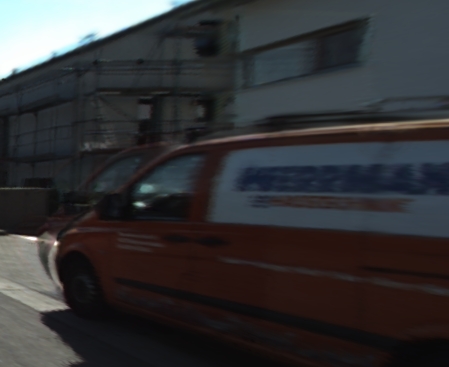}
         \caption{w/o occlusion check $\bv_k$}
        \label{fig: ablation occlusion}
     \end{subfigure} \\
     \begin{subfigure}[h]{0.45\linewidth}
         \centering
         \includegraphics[width=\textwidth]{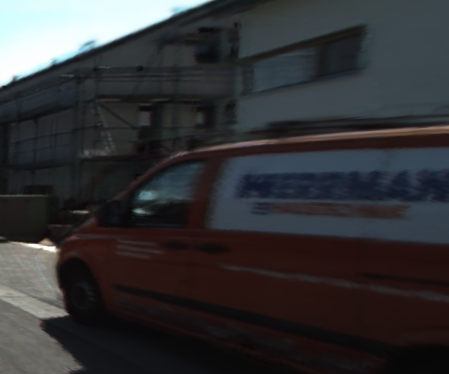}
        \caption{Full model}
        \label{fig:Full model}
     \end{subfigure}
      \begin{subfigure}[h]{0.45\linewidth}
         \centering
         \includegraphics[width=\textwidth]{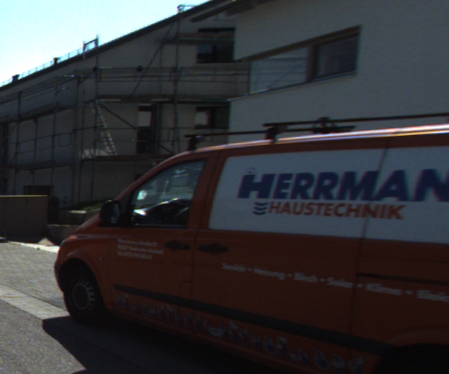}
        \caption{Ground Truth}
        \label{fig:ground truth}
     \end{subfigure}
     \vspace{-0.2cm}
             \caption{\textbf{Qualitative Results of Ablation Study.} We represent rendering images via feed-forward inference on one novel scene.}
        \label{fig:Ablation study.}
        \vspace{-0.4cm}
\end{figure}

For fast ablation experiments, We select a subset of the training sequences from the KITTI-360 datasets for training. We use these pretrained model to evaluate their feed-forward performance on drop50\% sparsity level to ablate the design choices of~\method.  \cref{tab:ablation} and \cref{fig:Ablation study.} presents the quantitative and qualitative results respectively.

\boldparagraph{Effectiveness of IBR Module}
We evaluate the impact of the IBR module by comparing our method to a variant that uses the initial point cloud's color without an appearance learning scheme. In this variant, global points learn geometric attributes via networks but retain appearance information from the initial point cloud. 
Due to the misalignment that exists in the monocular depth predictions, this na\"ive strategy results in blurry renderings in new scenes (see \cref{fig:wo ibr}). In contrast, our method leverages the IBR module, enabling high-detail rendering by predicting and retrieving colors from reference images. 

\boldparagraph{Effectiveness of Refine offset}
We now ablate the position refinement strategy.  Our model updates the primitive locations during training considering the predicted depth may be inaccurate. We compare our method with a variant that does not output the position offset $\Delta \mu_i$, using only the predicted positions as Gaussian locations. The results in \cref{tab:ablation} show a slight decrease in rendering quality, with more noticeable declines in SSIM and LPIPS metrics. This indicates that our generalized priors effectively compensate for depth estimation errors, facilitating improved rendering quality.

\boldparagraph{Effectiveness of Occlusion Check}
To investigate the occlusion check importance, we perform a study that removes the visibility maps and aggregates all 2D retrieved colors. As illustrated in \cref{fig: ablation occlusion}, the rendering quality in occluded regions suffers from noticeable blurriness. This is because 3D Gaussians can retrieve inconsistent colors due to the occlusions. By incorporating a visibility map $\bv_k$ as guidance, our model can effectively remove the impact of these inconsistencies, resulting in an overall improvement of about 0.3 dB PSNR as reported in \cref{tab:ablation}.

\begin{table}[t]
\centering
\small
\setlength{\tabcolsep}{20pt}
\begin{tabular}{@{}lccc@{}} 
\toprule
& PSNR$\uparrow$  & SSIM$\uparrow$  & LPIPS$\downarrow$ \\
\cline{2-4}
$W=1$ &22.54 &0.771 &0.216  \\
$W=3$ &22.83 &0.786  &0.190  \\
$W=5$  &22.84  &0.785 &0.201   \\
\bottomrule
\end{tabular}
\vspace{-0.2cm}
\caption{\textbf{Ablation Study} on different widow size.}
\vspace{-0.2cm}
\label{tab:color_window}
\end{table}

\boldparagraph{Analysis of color window size}
We further study the effect of different projection window. As presented in \cref{tab:color_window}, we compare three  window sizes: $W=\{1,3,5\}$ and evaluate their performance in feed-forward inference on novel scenes.  When $W=1$, each Gaussian retrieves a single pixel in each reference image for color prediction, resulting in the lowest PSNR. This is because inaccurate Gaussian positions lead to inaccurate 2D projections. With increased $W=3$, more color information is integrated to compensate for projection inaccuracies, resulting the better image quality. At $W=5$, while more information is queried, the redundancy can cause the image to become blurry, leading to an increase in LPIPS and higher memory consumption. We show more qualitative results in supplementary.

\section{Conclusion}
\label{sec:conclusion}
This paper presents~\method, a method for efficient urban scene reconstruction in a feed-forward manner. Unlike previous pixel-aligned 3DGS frameworks, we use geometric priors to construct a global volume and predict a standalone 3D representation, achieving state-of-the-art performance across several street-view datasets and enabling real-time rendering, making it well-suited for urban scenes. However, driving scenes still pose significant challenges, particularly in handling dynamic agents such as other vehicles. In future work, we will explore ways to generalize the reconstruction of dynamic street scenes.

\section*{Acknowledgements}
 We would like to thank Dzmitry Tsishkou and Sicheng Li for his useful suggestions and tips. This work is supported by NSFC under grant 62441223, 62202418, and U21B2004.

{\small
\bibliographystyle{ieeenat_fullname}
\bibliography{11_references}
}

\clearpage
\setcounter{page}{1}
\maketitlesupplementary

\section{Analysis of the convergence of $\Delta \mu_i$}
In this section, we will show the position offset $\Delta \mu_i$ converge to a fixed value at the infinite training iterations. Recall that the  $\Delta \mu_i$ in the iteration $k$ can be formulated as:
\begin{align}
   \bff_i = & \bF (\mu_i^{init} + \Delta \mu_i^{k-1}) \label{eq:offset4} \\
   \Delta \mu_i^{k} = & \textit{Tanh}(\mathcal{D}_{pos}(\bff_i))\cdot v_\text{size}  \label{eq:offset3}
\end{align}
We can derive the $\Delta \mu_i^{k}$ as:
\begin{align}
   \Delta \mu_i^{k} = & \textit{Tanh}(\mathcal{D}_{pos}(\bF (\mu_i^{init} + \Delta \mu_i^{k-1})))\cdot v_\text{size}
\end{align}
We define
$\mathcal{H}\triangleq \textit{Tanh}(\mathcal{D}_{pos}(\bF(\cdot)))\cdot v_\text{size}$ for simplicity, i.e.,
\begin{equation}
   \Delta \mu_i^{k} = \mathcal{H}(\mu_i^{init} + \Delta \mu_i^{k-1})
   \label{eq5:taylor}
\end{equation}
Given that $ \Delta \mu_i$ small, we apply the first-order Taylor expansion to approximate Eq.~\ref{eq5:taylor}:
\begin{equation}
   \Delta \mu_i^{k} =  \mathcal{H}(\mu_i^{init}) +  \cdot \left. \frac{\partial  \mathcal{H}}{\partial \mu}\right|_{\mu_i^{init}} \Delta \mu^{k-1}_i
   \label{eq5:taylor3}
\end{equation}
Considering that $\boldsymbol{\beta}\triangleq  \mathcal{H}(\mu_i^{init})$ and $ \boldsymbol{\Gamma}\triangleq  \left. \frac{\partial  \mathcal{H}}{\partial \mu}\right|_{\mu^{init}}$  are both constants, we reformulate   Eq.~\ref{eq5:taylor3} as:
\begin{align}
   \Delta \mu_i^{k} = & \boldsymbol{\beta} + \boldsymbol{\Gamma} \Delta \mu^{k-1}_i
   \label{eq5:taylor4}
\end{align}
The training converges in our experimental observations. Therefore, we assume $\Delta \mu^{\infty}_i$ converges as $k$ approaches the infinite step, i.e., $k \rightarrow \infty$. This allows us to derive $\Delta \mu^{\infty}_i$:
\begin{align}
    \Delta \mu^{\infty}_i=&\boldsymbol{\boldsymbol{\beta}}+\boldsymbol{\boldsymbol{\Gamma}} \Delta \mu^{\infty}_i\\
    =&(\boldsymbol{I}-\boldsymbol{\boldsymbol{\boldsymbol{\Gamma}}})^{-1} \boldsymbol{\boldsymbol{\beta}}
\end{align}
This indicates that $\mu^{\infty}_i$ converges to a constant value. In supplementary \cref{sec:delta_mu}, we show that $\mu_i^t$ converges quickly at early steps during inference.

\section{Implementation Details}
\subsection{Remove Depth Outliers}
We begin by applying a depth consistency check to filter out noisy depth data. Specifically, we unproject the depth map $D_i$ of $i$th frame to 3D and reproject it to a nearby view $j$, obtaining a projected depth $D_{i\rightarrow j}$. Next, we compare $D_{i\rightarrow j}$ and $D_j$ and filter out depths where the absolute relative error exceeds an empirical threshold $\sigma = 0.2m$. We formulate this process as:

\begin{equation}
M_d=\left|D_{i\rightarrow j}-D_j\right|<\sigma, D_{i\rightarrow j}=D_{j}\left(\pi_{j} \pi_i^{-1}\left(\mathbf{u}_i\right)\right)
\end{equation}
where $\mathbf{u}_i$ denotes pixel coordinates of $i$th frame and $M_d$ represents the geometric consistency mask. We further utilize the 3D statistical filter in Open3D~\cite{zhou2018open3d} to remove the clear floaters in the unprojected point clouds for each frame. In our implementation, we set the number of neighbors to 20 and the standard deviation ratio to 2.0.

\subsection{ Foreground Gaussian Details}
Following~\cite{3dgs}, $\boldsymbol{\Sigma}$ is decomposed into two learnable components: rotation matrix $\mathbf{R}$ and a scaling matrix $\mathbf{S}$ to holds practical physical significance, see the following formula:
\begin{equation}
\boldsymbol{\Sigma}=\mathbf{R} \mathbf{S} \mathbf{S}^T \mathbf{R}^T
\end{equation}
To allow independent optimization of both factors, we use a 3D vector $s$ representing scaling and a quaternion $q$ for rotation separately. Instead of directly learning scales $s$, we initialize the $s_\text{init}$ with the average distance of $K$ nearest neighbors using the KNN algorithm and learn the scales residual $\Delta s$ from the volume latent feature $\bff_i$ via a decoder $\mathcal{D}_{cov}$. We experimentally observe that the residual learning strategy helps our network converge faster and enhances the model capacity. 
\begin{equation}
\mathbf{s} = \mathbf{s}_\text{init} + \Delta \mathbf{s}, ~~~ \Delta \mathbf{s}=\mathcal{D}_{cov}(\bff_i)
\end{equation}

\subsection{ Hemisphere Background Details}
We also develop the generalizable hemisphere model for the background which typically lies hundreds of meters away from the vehicle, as discussed in the main paper. Specifically, we initialize the background as a hemisphere with a fixed radius $r_{bg}=100m$ as a hyperparameter, with its center positioned at the midpoint of the foreground volume. This background hemisphere moves along with the vehicle such that the relative distance is fixed wrt. the target camera. We project the points onto $K$ reference images to retrieve their 2D color $\{\bc_{k}\}_{k=1}^K$ to regress all gaussian parameters through network $\mathcal{M}_{bg}$ as mentioned in the main paper. Similar to the foreground framework, we initialize the Gaussian scales using the KNN algorithm and learn their scale residuals $\Delta \mathbf{s}_{bg}$ for each Gaussian.
\begin{equation}
\mathbf{s}_{b g}= \Delta \mathbf{s}_{bg} + \mathbf{s}_{b g}^{init}, ~~~ \Delta \mathbf{s}_{bg}= \mathcal{M}_{b g}\left(\left\{\mathbf{c}_k\right\}_{k=1}^K\right)
\end{equation}
 For opacity and rotation, we explicitly set the opacity to 1 and the rotation to a unit quaternion $[1,0,0,0]$ as a reasonable initialization, without involving them in the network optimization.

\subsection{SparseCNN Network Architecture}
We build a generalizable efficient 3DCNN $\psi^\text{3D}$ to provide the geometric priors for the foreground contents.  Given the global point cloud $\mathcal{P} \in \mathbb{R}^{N_p \times 3}$, we quantize the point cloud with the voxel size $v_\text{size} = 0.1m$ and fed these sparse tensors into the $\psi^\text{3D}$ to predict the latent feature volume$\bF$. The sparse 3DCNN uses a U-Net like architecture with skip connections, comprising some convolution and transposed convolution layers.  The details of 3DCNN are listed in the \cref{tab:sparse cnn}. We use torchsparse as the implementation of $\psi^\text{3D}$.

\begin{table}[h]
  \centering
    \begin{tabular}{@{}l|cc@{}}
    \toprule
    \multicolumn{3}{l}{$\textbf{SparseCNN Network Architecture}$} \\
    \hline
    \textbf{Layer} &\textbf{Description} &\textbf{In/Out Ch.}  \\
    Conv$_{i=0}$ & kernel = $3\times3\times3$, \text{stride} = $1$ & 3/16  \\
    Conv$_{i=1}$ & kernel = $3\times3\times3$, \text{stride} = $2$ & 16/16  \\
    Conv$_{i=2}$ & kernel = $3\times3\times3$, \text{stride} = $1$ & 16/16  \\
    Conv$_{i=3}$ & kernel = $3\times3\times3$, \text{stride} = $2$ & 16/32  \\
    Conv$_{i=4}$ & kernel = $3\times3\times3$, \text{stride} = $1$ & 32/32  \\
    Conv$_{i=5}$ & kernel = $3\times3\times3$, \text{stride} = $2$ & 32/64  \\
    Conv$_{i=6}$ & kernel = $3\times3\times3$, \text{stride} = $1$ & 64/64  \\
    DeConv$_{i=7}$ & kernel = $3\times3\times3$, \text{stride} = $2$ & 64/32  \\
    DeConv$_{i=8}$ & kernel = $3\times3\times3$, \text{stride} = $2$ & 32/16  \\
    DeConv$_{i=9}$ & kernel = $3\times3\times3$, \text{stride} = $2$ & 16/16  \\
    
     \bottomrule
    \end{tabular}%
    \vspace{0.1cm}
    \caption{\textbf{Architecture of SparseConvNet}. Each layer consists of sparse convolution, batch normalization, and ReLU.} 
  \label{tab:sparse cnn}%
\end{table}%

\section{Baselines}
In this section, we discuss the state-of-the-art baselines used for comparison with our approach.

\boldparagraph{Feed-Forward NeRFs}
We adopt the official implementations of MVSNeRF~\cite{mvsnerf}, MuRF~\cite{xu2023murf} and EDUS~\cite{miao2024edus}. For each method, we retrain the model using 160 sequences from KITTI-360~\cite{liao2022kitti} under a 50\% drop rate. We select the three nearest training frames of the target view as reference images for these methods. MVSNeRF and MuRF utilize multi-view stereo (MVS) algorithms to construct the cost volume and apply 3DCNN to reconstruct a neural field while EDUS leverages the depth priors to learn a generalizable scene representation.

\boldparagraph{Feed-Forward 3DGS}
We adopt the official implementations of PixelSplat~\cite{charatan2024pixelsplat} and MVSplat~\cite{chen2024mvsplat}. We find that using three reference images caused color shifts at novel viewpoints in urban scenes, so we use the two nearest training frames to achieve optimal performance following the original paper. PixelSplat predicts 3D Gaussians with a two-view epipolar transformer and then spawns per-pixel Gaussians. MVSplat exploits multi-view correspondence information for geometry learning and predicts 3D Gaussians from image features.  Both methods are trained on a single Nvidia RTX V100 using the full resolution of KITTI-360.

Additionally, as we illustrated in the teaser figure in the main paper, these pixel-align 3DGS methods predict inconsistent 3DGS when accumulating multiple local volumes. Note that to ensure a fair comparison, we conduct experiments using a single local volume following their default setting(use 2 reference images), as reported in Table 1 and Figure 4 in the main paper.

\boldparagraph{Test-Time Optimization Methods}
To evaluate our fine-tuning results, we compare them against recent test-time optimization methods under the 50\% drop rate. Specifically, we use the latest version of Nerfacto~\cite{tancik2023nerfstudio} provided by Nerfstudio and the official codebase of 3DGS~\cite{3dgs}. Nerfacto is a combination of many published methods that demonstrate strong performance on real-world data, including pose refinement, appearance embedding, scene contraction, and hash encoding. For 3DGS, we initialize the 3DGS model with our global point cloud to ensure a fair comparison.

\section{Additional Experimental Results}
\subsection{Monocular Depth Modularity}
To further evaluate the sensitivity of our method to different depth estimation approaches, we conduct experiments using two distinct metric depth estimators: Metric3D~\cite{yin2023metric3d} and UniDepth~\cite{piccinelli2024unidepth}.  Specifically, we pretrain our model with Metric3D, and evaluate using depth maps of two different models for  feed-forward inference on novel scenes. As shown in \cref{tab:depth sensitivity}, our~\method~consistently outperforms the baseline methods on both depth estimators, demonstrating its robustness in handling depth predictions across varying distributions.

\begin{table}[h]
\centering
\small
\setlength{\tabcolsep}{12pt}
\begin{tabular}{cccc} 
\toprule 
Depth Method & PSNR$\uparrow$  & SSIM$\uparrow$  & LPIPS$\downarrow$ \\
\cline{1-4}
Metric3D~\cite{yin2023metric3d} &24.43 &0.786 &0.202  \\
UniDepth~\cite{piccinelli2024unidepth} &23.38 &0.775  &0.223  \\
\bottomrule
\end{tabular}
\caption{\textbf{Depth Sensitivity Experiment} The results are averaged on five testsets from the Waymo dataset.}
\vspace{-0.2cm}
\label{tab:depth sensitivity}
\end{table}

\subsection{Additional Ablation for $\Delta \mu_i$}
\label{sec:delta_mu}
As mentioned in the main paper, $\Delta \mu_i$ depends on the previous estimation $\Delta \mu^{prev}_i$, but it stabilizes at a stationary point after infinite training iterations. Note that $\mathcal{D}_{pos}$ is designed to continually decode the offset $\Delta$ wrt. $\mu_{init}$, even at the ideal location, avoiding an infinite loop caused by toggling between the ideal offset and zero. We further conduct experiments by recursively updating the offsets ($i=0,1,2,3$) during inference to verify its convergence. As reported in \cref{tab:recursion offset}, our pretrained model successfully refines the noisy primitive's position after the first inference (increasing PSNR by approximately 0.46 dB) and maintains a stable value of 23.78dB even with additional updates.

\begin{table*}[t]
\centering
\small
\setlength{\tabcolsep}{11pt}
\begin{tabular}{@{}l|ccc|ccc@{}}
\toprule
\multirow{2}{*}{Method} & \multicolumn{3}{c|}{Waymo} & \multicolumn{3}{c}{KITTI-$360$ }\\
             & PSNR(dB)$\uparrow$  & SSIM$\uparrow$ & LPIPS$\downarrow$ & PSNR(dB)$\downarrow$ & SSIM$\uparrow$    &  LPIPS$\downarrow$    \\ \midrule
MuRF~\cite{xu2023murf}     &23.66 &0.746  &0.256 &19.83 &0.669 &0.340  \\
EDUS~\cite{miao2024edus}    & 23.41  & 0.769  & \textbf{0.147} &20.13 &0.659  &0.257     \\ 
MVSplat~\cite{chen2024mvsplat}    &24.08  &0.758 &0.197 &17.80  &0.581  &0.361  \\
Ours          &\textbf{25.06}  &\textbf{0.820} &0.189 &\textbf{21.23}  &\textbf{0.738}  &\textbf{0.222}  \\
\bottomrule
\end{tabular}
\vspace{-0.2cm}
\caption{\textbf{Quantitative results on Waymo and KITTI-360 datasets} with other generalizable methods. All models are trained on the Waymo dataset using drop50\% sparsity level. Metrics are averaged on five validation scenes without any finetuning.}
\label{tab:feed forward_waymo}
\end{table*}

\begin{table}[h]
\centering
\small
\setlength{\tabcolsep}{20pt}
\begin{tabular}{@{}lccc@{}} 
\toprule
& PSNR$\uparrow$  & SSIM$\uparrow$  & LPIPS$\downarrow$ \\
\cline{2-4}
$i=0$ &23.329 &0.794 &0.175  \\
$i=1$ &23.787 &0.819  &0.171  \\
$i=2$  &23.778  &0.819 &0.171   \\
$i=3$  &23.786  &0.819 &0.171   \\
\bottomrule
\end{tabular}
\vspace{-0.2cm}
\caption{\textbf{Ablation Study} on recursion of $\Delta \mu_i$}
\vspace{-0.2cm}
\label{tab:recursion offset}
\end{table}

\begin{table}[h]
\centering
\captionsetup{skip=2pt}
{
\begin{tabular}{@{}ccccc@{}}
\toprule 
Layers & Width & PSNR$\uparrow$ & SSIM$\uparrow$ & LPIPS$\downarrow$\\
\midrule
2 & 64 & 22.61 & 0.780 & 0.192 \\
3 & 128 & 22.69 & 0.785 & 0.189  \\
4 & 128 & 22.60 & 0.785 & 0.191  \\
\bottomrule
\end{tabular}
}
\caption{\textbf{Ablation study} of the background MLP capacity.}
\label{tab: ablation background capacity}
\end{table}

\subsection{Training on Waymo}
To verify our method's performance given different training sequences, we train our method on the Waymo dataset and evaluate its feed-forward performance on Waymo and KITTI-360, as shown in the \cref{tab:feed forward_waymo}. Our method consistently achieves state-of-the-art performance in terms of PSNR and SSIM metrics, indicating its robustness for different driving data distributions.

\subsection{ Ablation Experiments for Background MLP}
Our background model primarily blends colors from nearby reference images rather than learning textures from scratch. As shown in \cref{tab: ablation background capacity}, increasing the layers and width of the background MLP does not yield significant improvements. A light-weight two-layer MLP provides sufficient capacity to reconstruct backgrounds while minimizing computational overhead.

\section{More Qualitative Results}
\subsection{More Qualitative Results in Ablation Study}

\begin{figure}[h]
  \centering
   \includegraphics[width=1.0\linewidth]{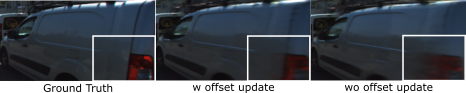}
   \caption{Qualitative Results of offset refine strategy}
   \label{fig: offset refine}
   \vspace{-0.5cm}
\end{figure}

\begin{figure}[h]
  \centering
   \includegraphics[width=1.0\linewidth]{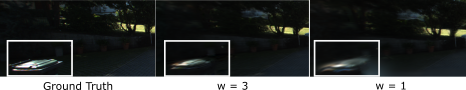}
   \caption{Qualitative Results of windows size strategy}
   \label{fig: window size}
   \vspace{-0.4cm}
\end{figure}

Removing offset refinement and occlusion check leads to visible artifacts in small regions, such as the car  in Fig. 7(b) and  \cref{fig: offset refine}. While these components don’t significantly affect overall quantitative results, they improve local visual quality. 
Similarly, the color projection window, which compensates for inaccurate Gaussian positions, also improves local visual quality, see \cref{fig: window size}.

\subsection{More Feed-Forward Inference Qualitative Results}
Our method enables efficient reconstruction and real-time photorealistic NVS from flexible sparse street view images. We provide more qualitative results on the KITTI-360  dataset via a feed-forward inference under drop50\% setting, as shown in \cref{fig:more res}. 

\section{Limitions}
\begin{figure*}[htbp]
     \centering
     \setlength{\tabcolsep}{1pt}
     \def\mywidth{.33}
     \def\reduceheight{-3.pt}
     \begin{tabular}{ccc}
     \includegraphics[width=\mywidth\linewidth]{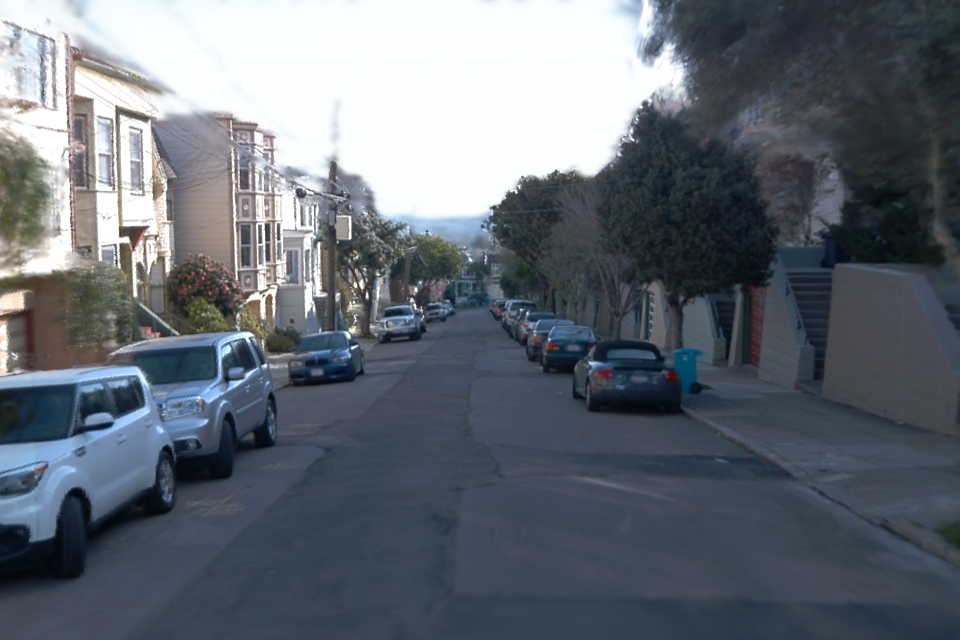} &
      \includegraphics[width=\mywidth\linewidth]{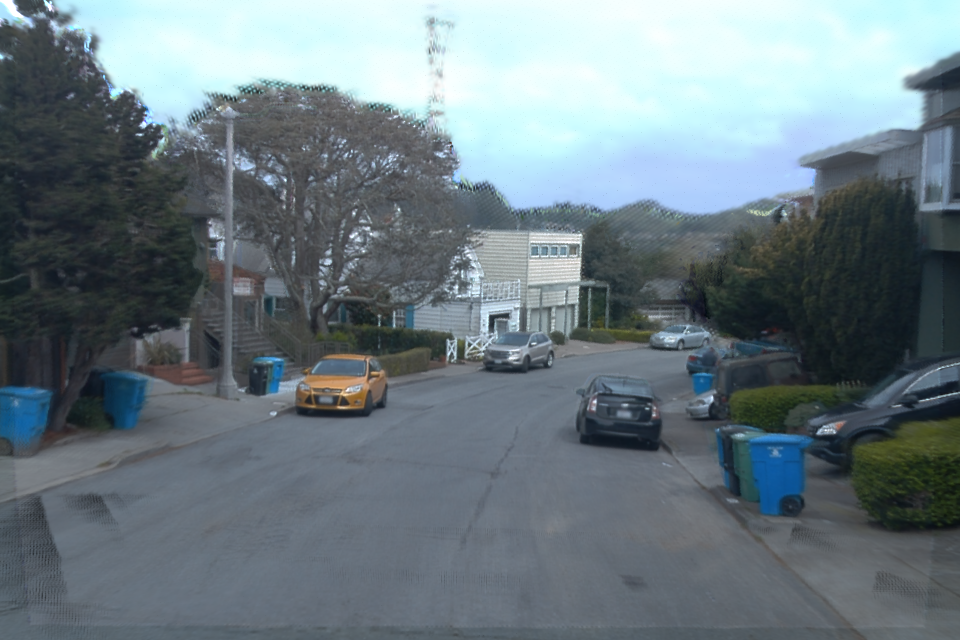} &
      \includegraphics[width=\mywidth\linewidth]{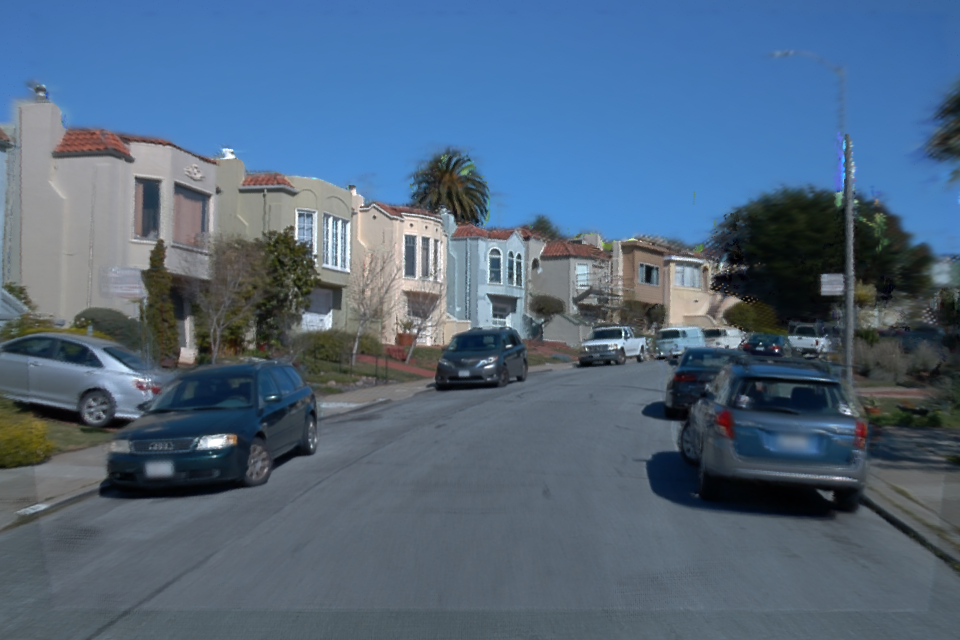} 
      \vspace{\reduceheight}\\

     \end{tabular}\vspace{-0.2cm}
     \caption{{\bf Limitations in thin structure and background}.}
     \label{fig:failure case}
     \vspace{-0.021cm}
    \end{figure*}

\begin{figure*}[htbp]
     \centering
     \setlength{\tabcolsep}{1pt}
     \def\mywidth{.33}
     \def\reduceheight{-3.pt}
     \begin{tabular}{ccc}
      \includegraphics[width=\mywidth\linewidth]{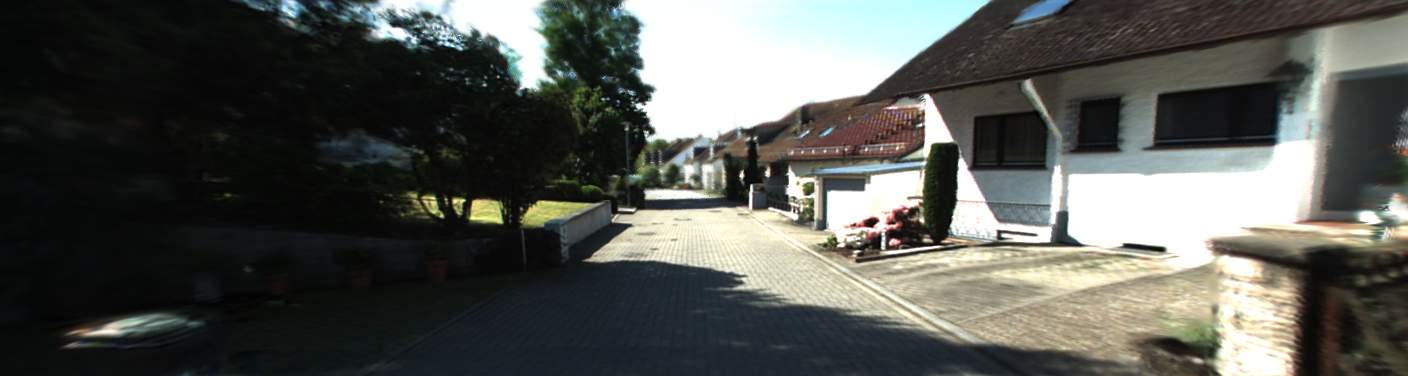} &
      \includegraphics[width=\mywidth\linewidth]{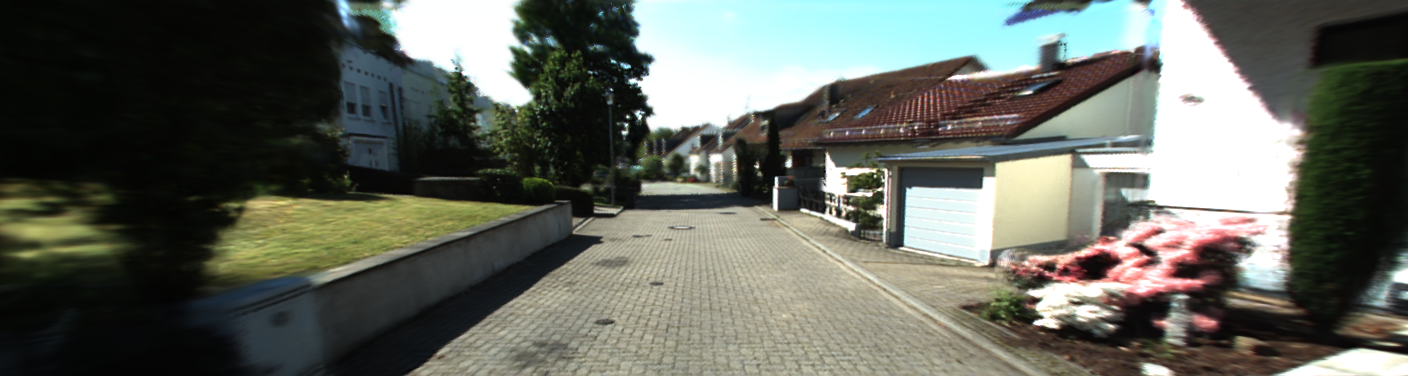} &
      \includegraphics[width=\mywidth\linewidth]{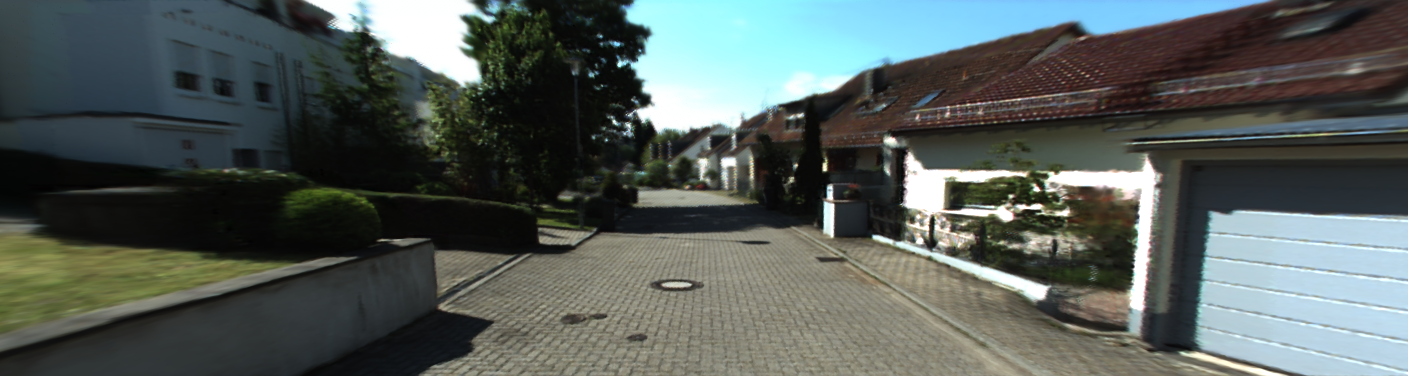} 
      \vspace{\reduceheight}\\
      \includegraphics[width=\mywidth\linewidth]{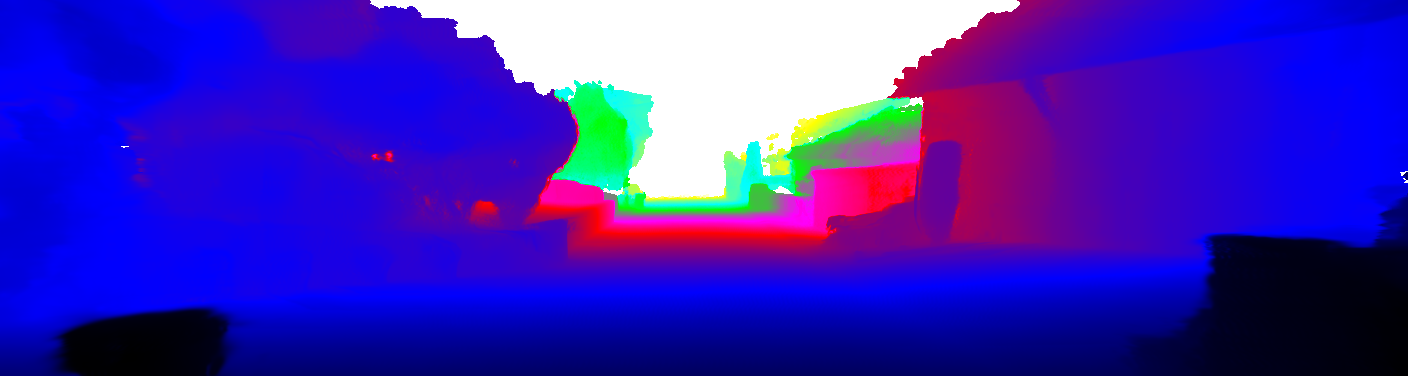} &
      \includegraphics[width=\mywidth\linewidth]{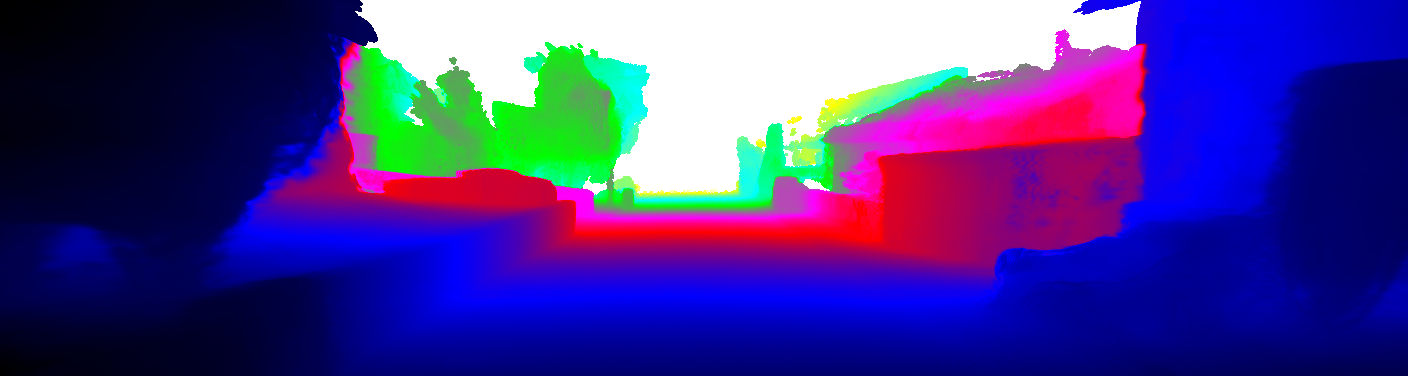} &
      \includegraphics[width=\mywidth\linewidth]{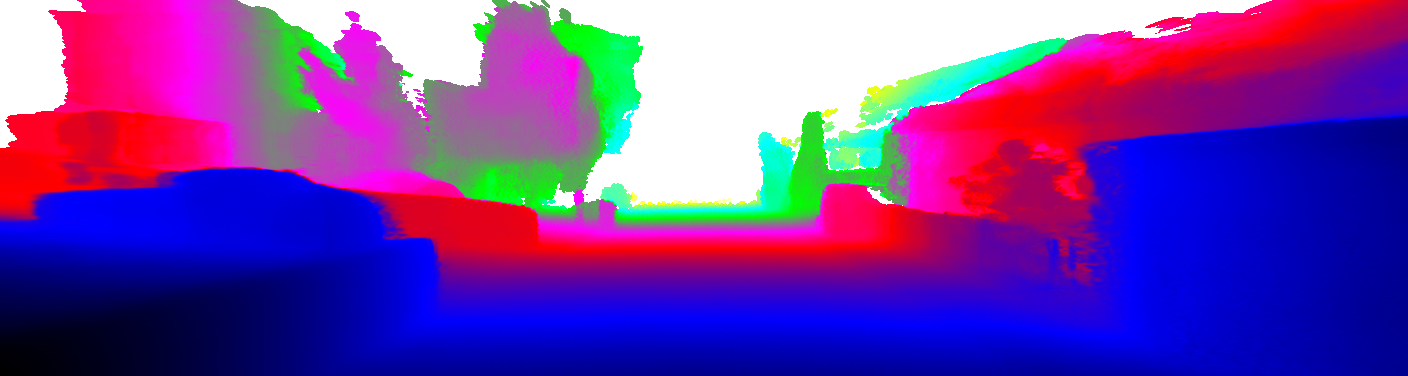} 
      \vspace{\reduceheight}\\
      
      \includegraphics[width=\mywidth\linewidth]{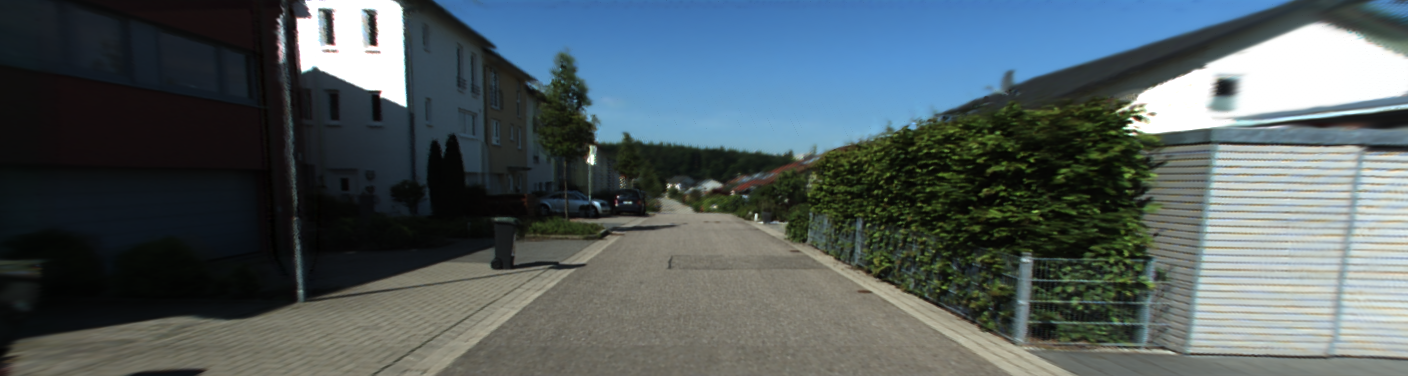} &
      \includegraphics[width=\mywidth\linewidth]{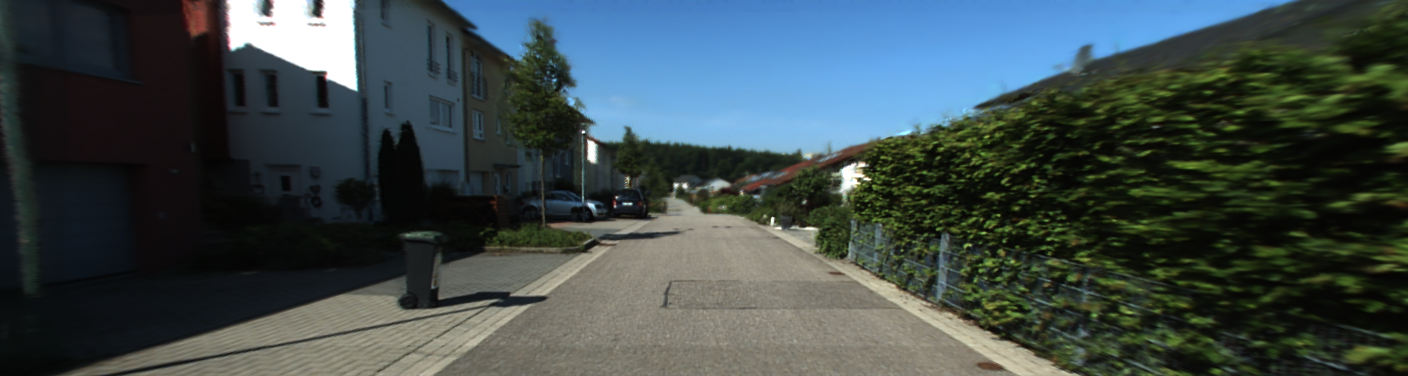} &
      \includegraphics[width=\mywidth\linewidth]{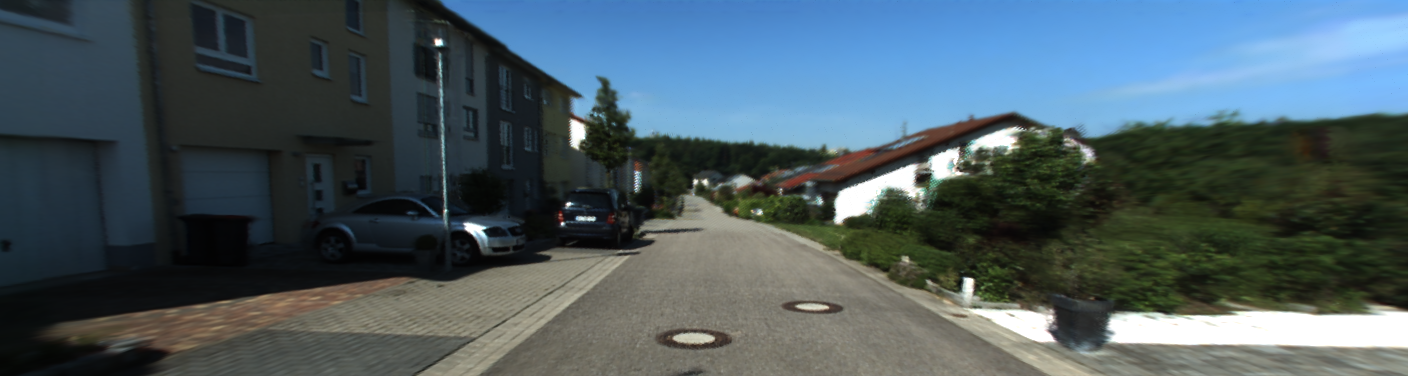} 
      \vspace{\reduceheight}\\
      \includegraphics[width=\mywidth\linewidth]{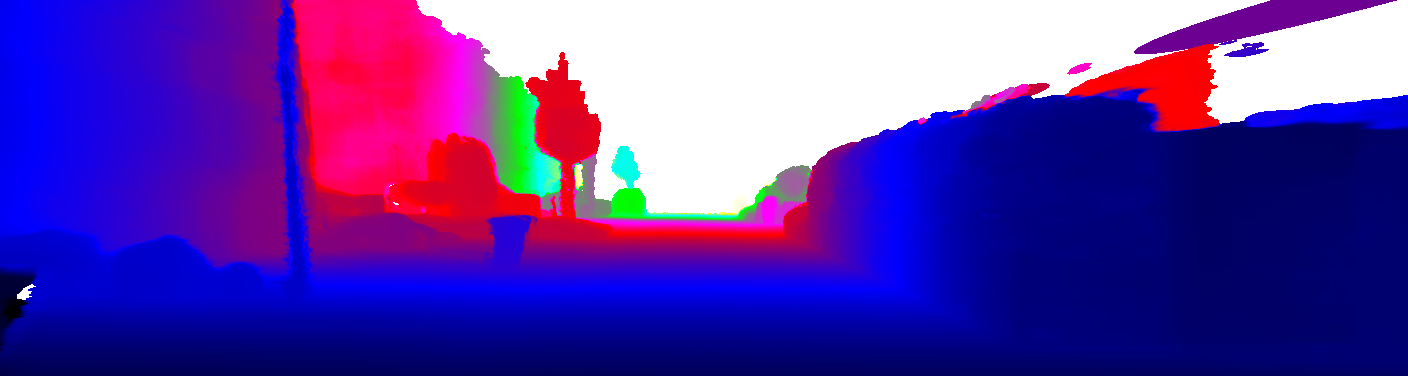} &
      \includegraphics[width=\mywidth\linewidth]{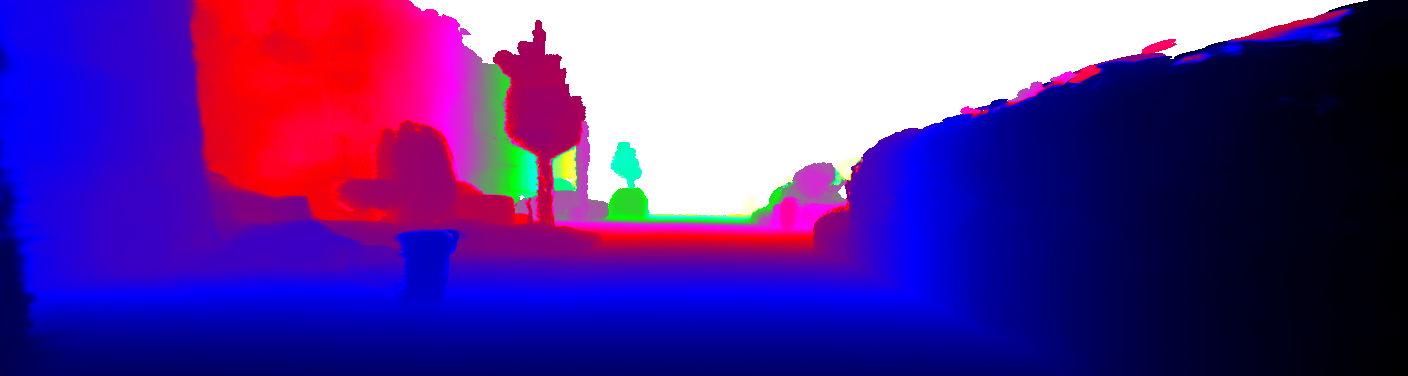} &
      \includegraphics[width=\mywidth\linewidth]{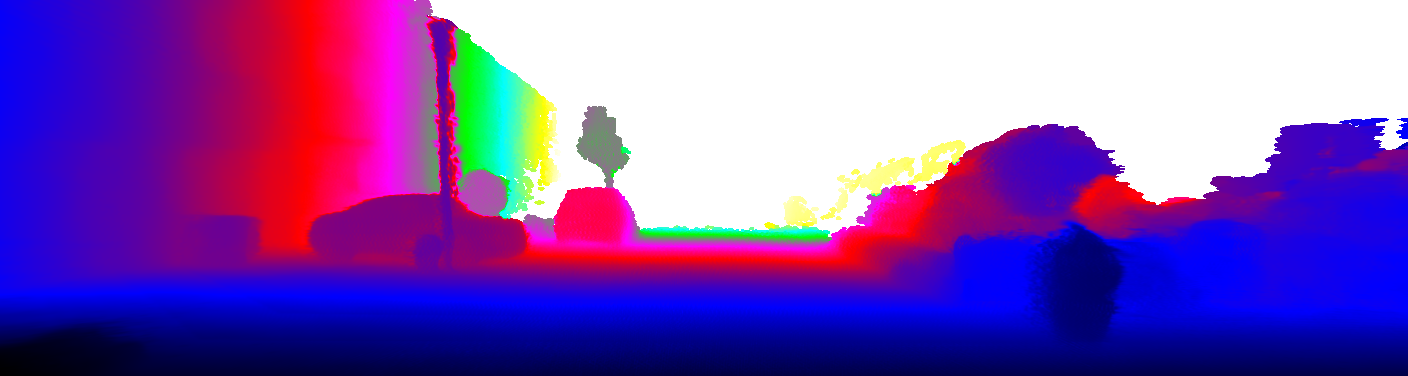} 
      \vspace{\reduceheight}\\
      
      \includegraphics[width=\mywidth\linewidth]{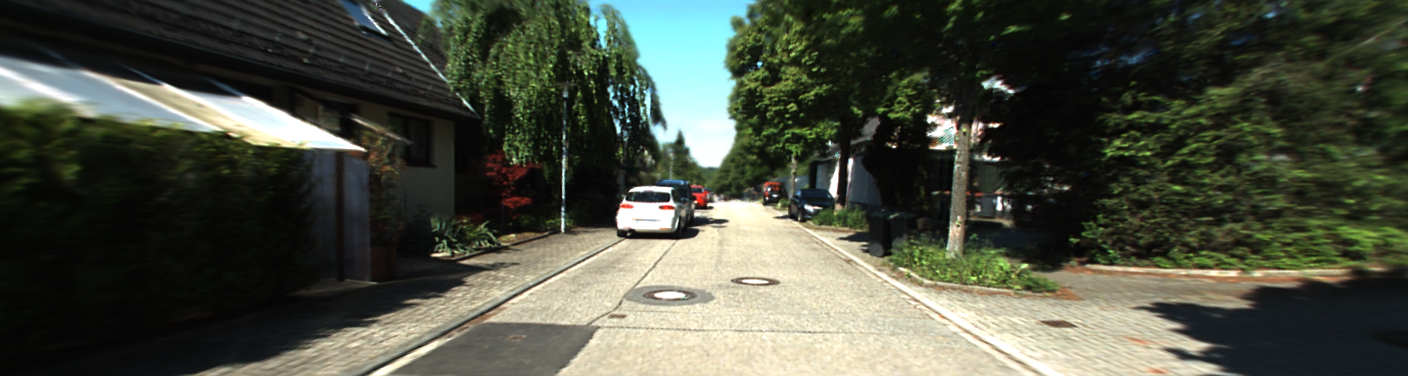} &
      \includegraphics[width=\mywidth\linewidth]{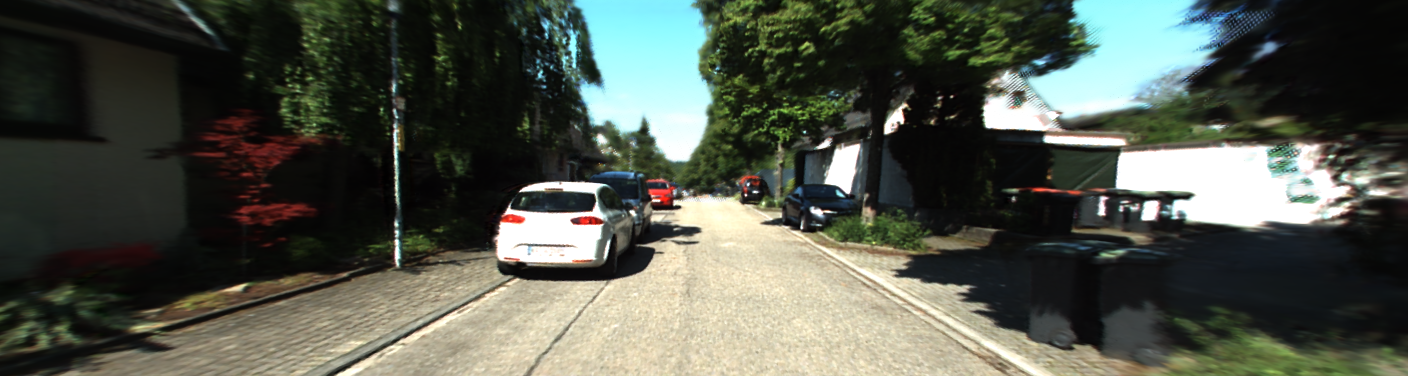} &
      \includegraphics[width=\mywidth\linewidth]{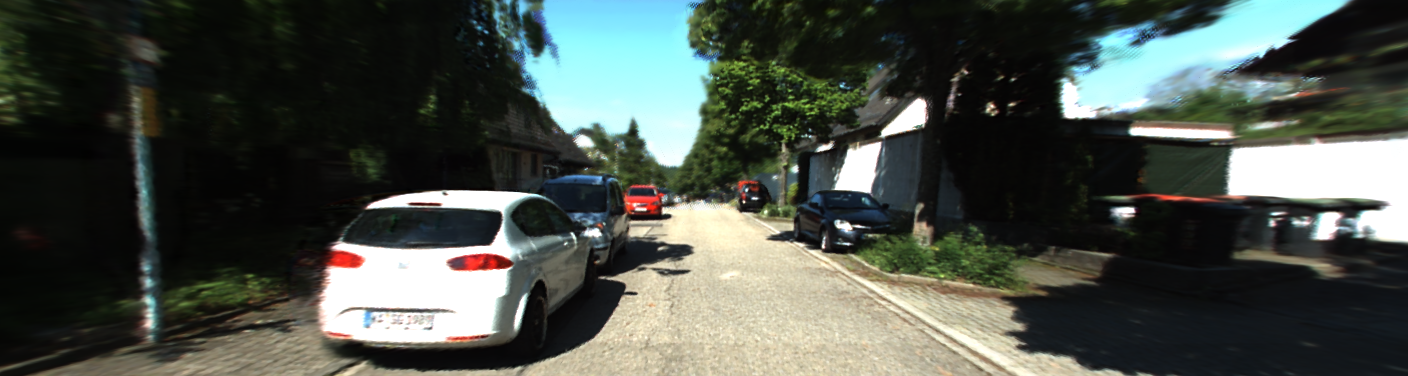} 
      \vspace{\reduceheight}\\
      \includegraphics[width=\mywidth\linewidth]{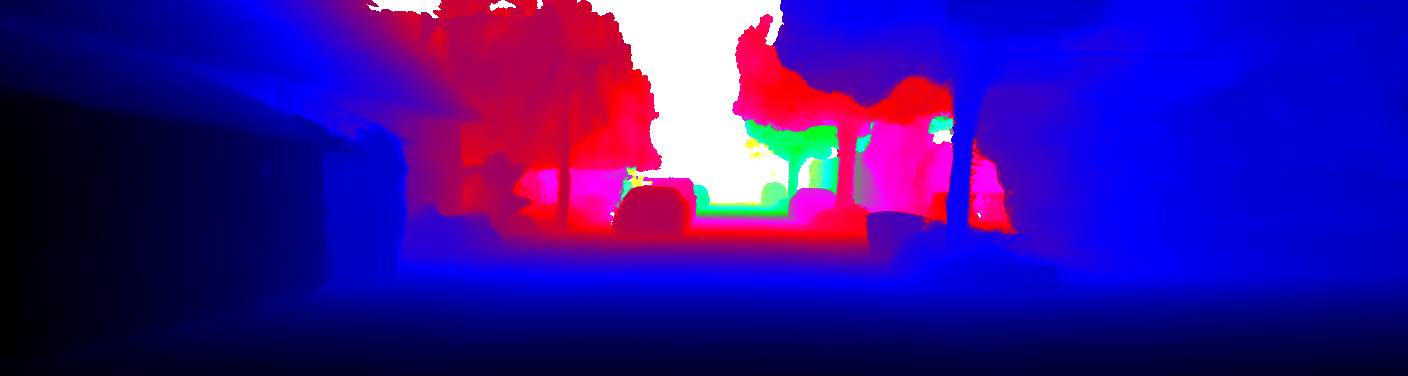} &
      \includegraphics[width=\mywidth\linewidth]{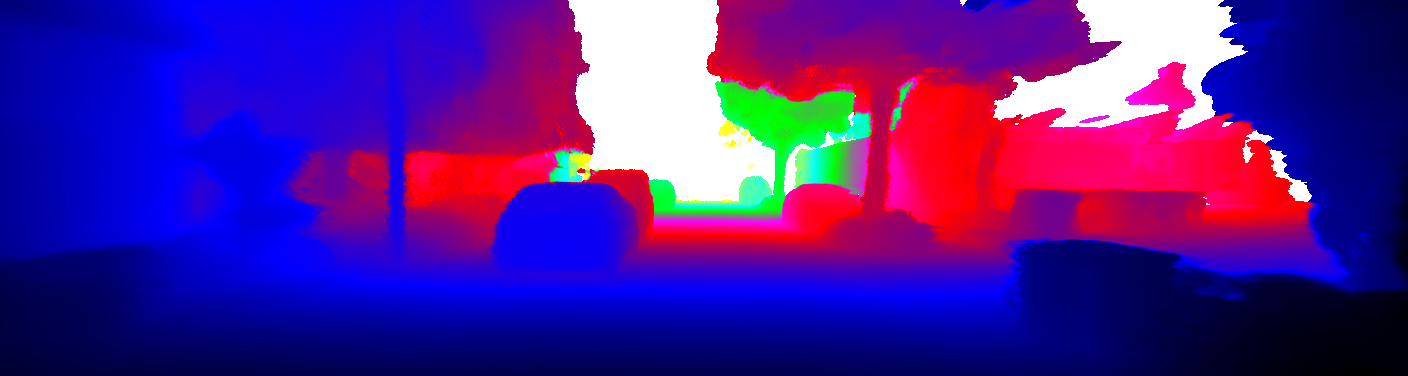} &
      \includegraphics[width=\mywidth\linewidth]{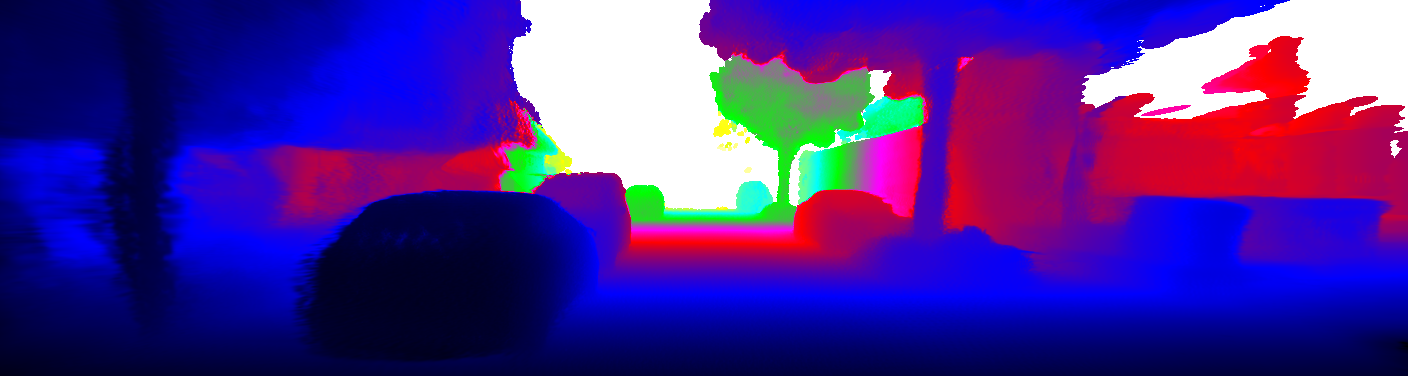} 
      \vspace{\reduceheight}\\

      \includegraphics[width=\mywidth\linewidth]{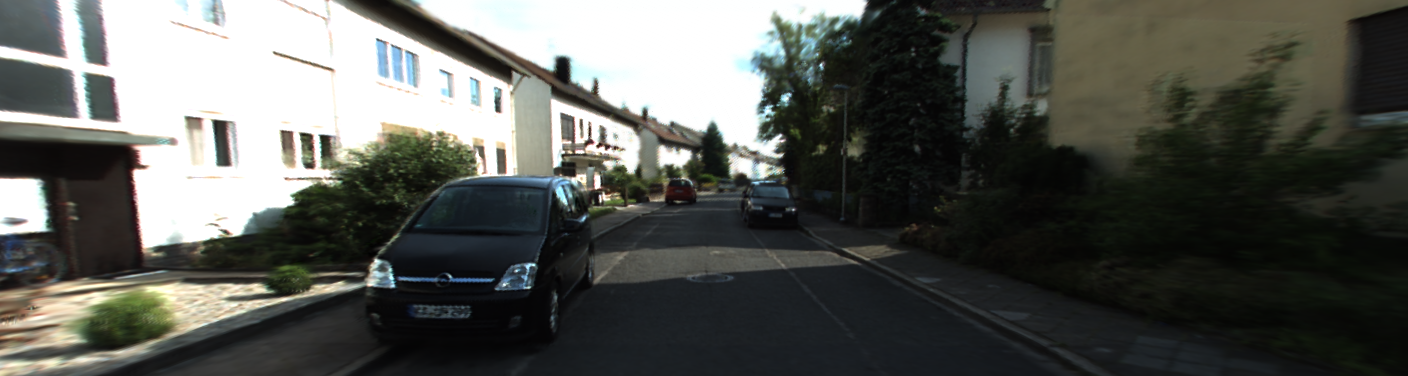} &
      \includegraphics[width=\mywidth\linewidth]{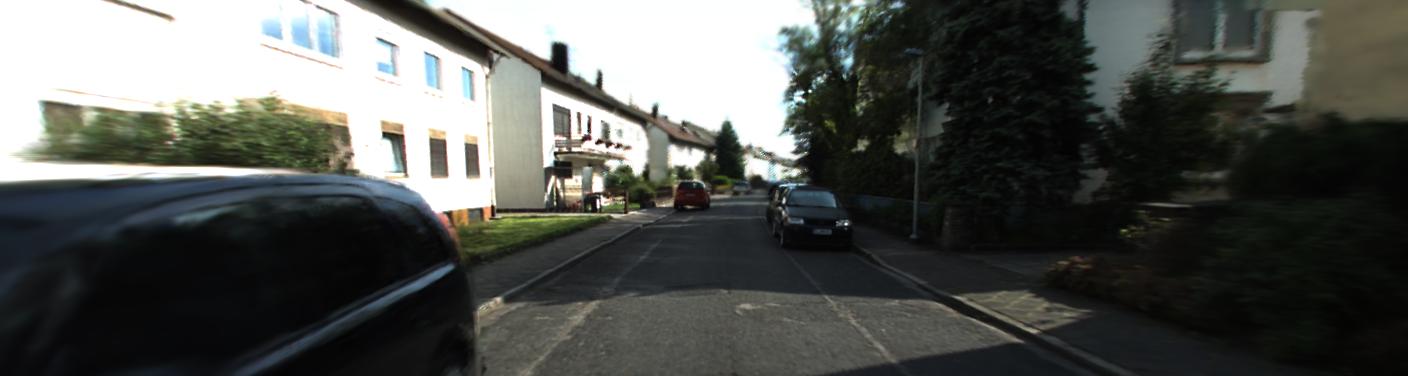} &
      \includegraphics[width=\mywidth\linewidth]{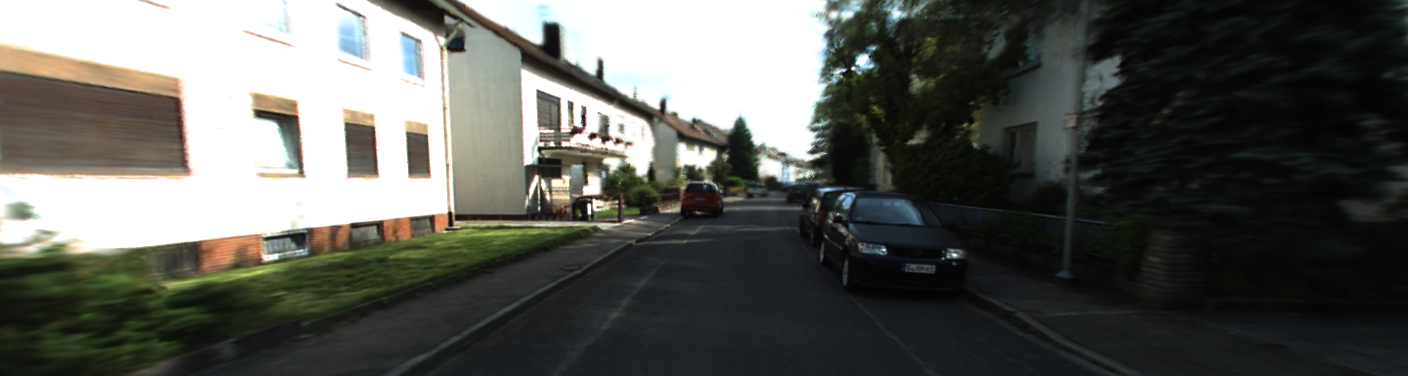} 
      \vspace{\reduceheight}\\
      \includegraphics[width=\mywidth\linewidth]{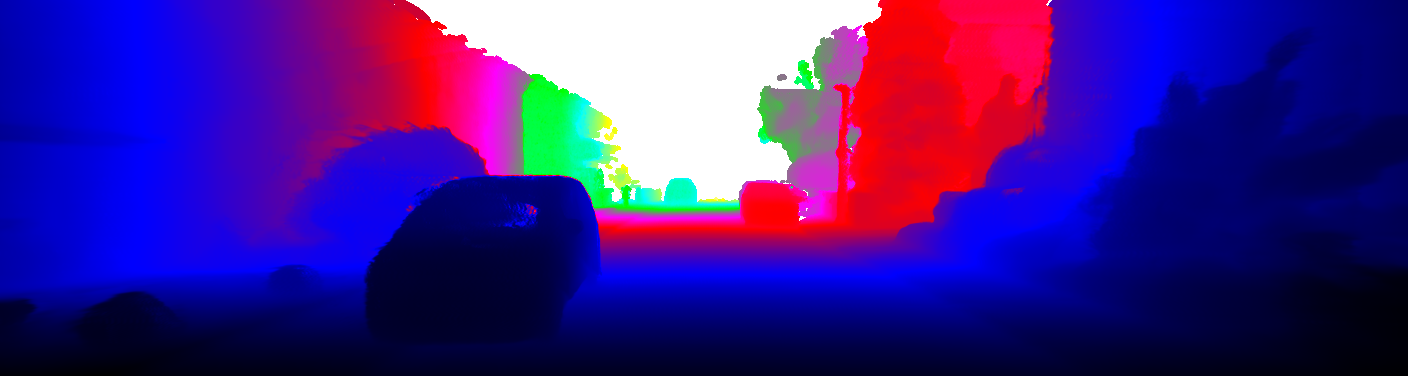} &
      \includegraphics[width=\mywidth\linewidth]{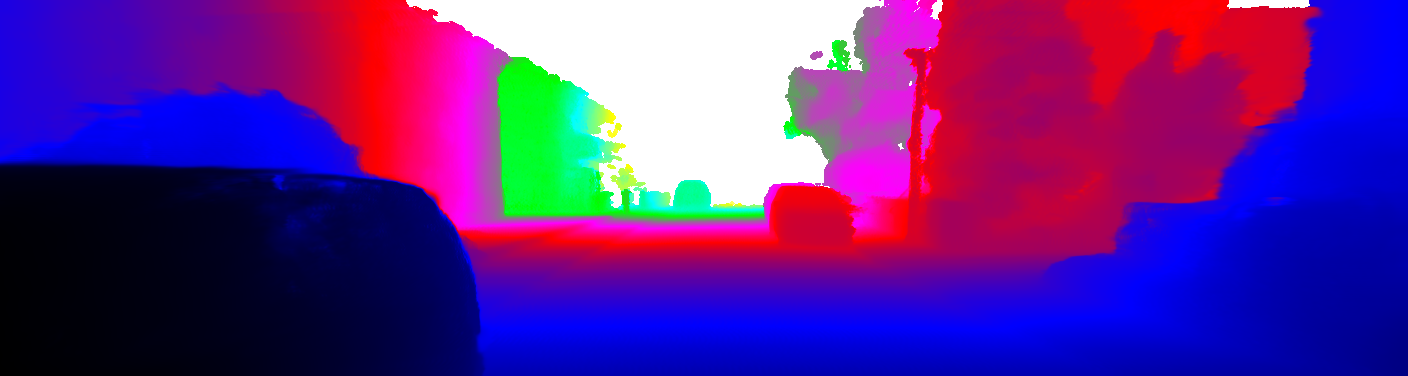} &
      \includegraphics[width=\mywidth\linewidth]{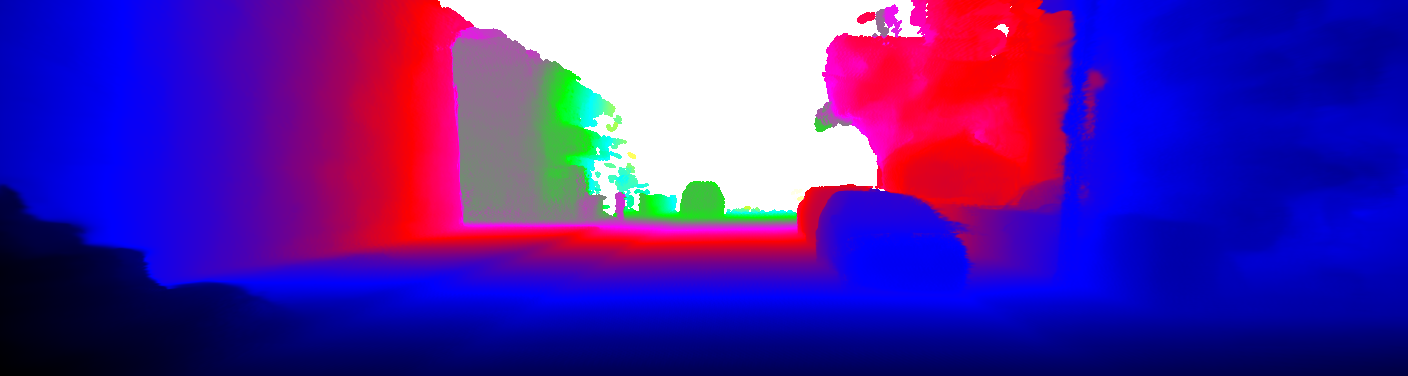} 
      \vspace{\reduceheight}\\

      \includegraphics[width=\mywidth\linewidth]{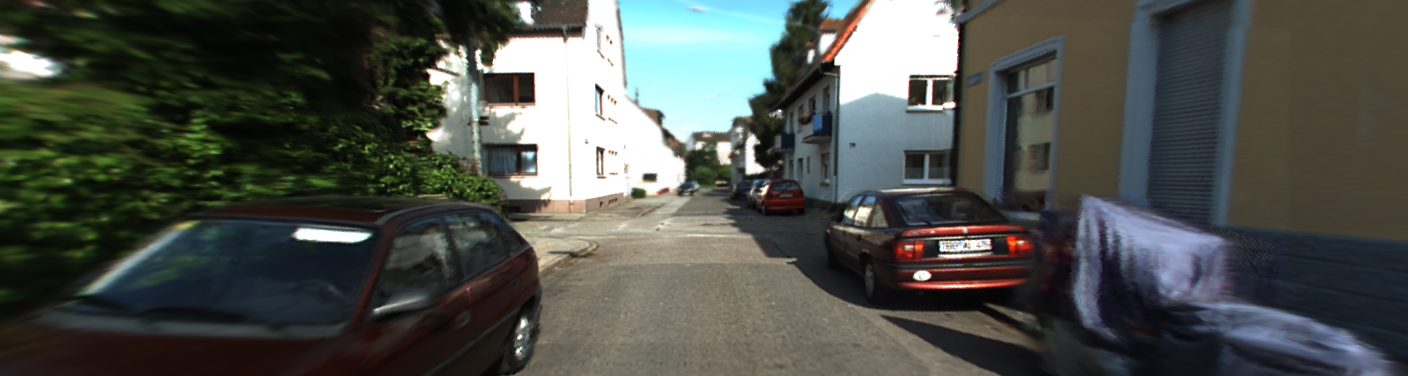} &
      \includegraphics[width=\mywidth\linewidth]{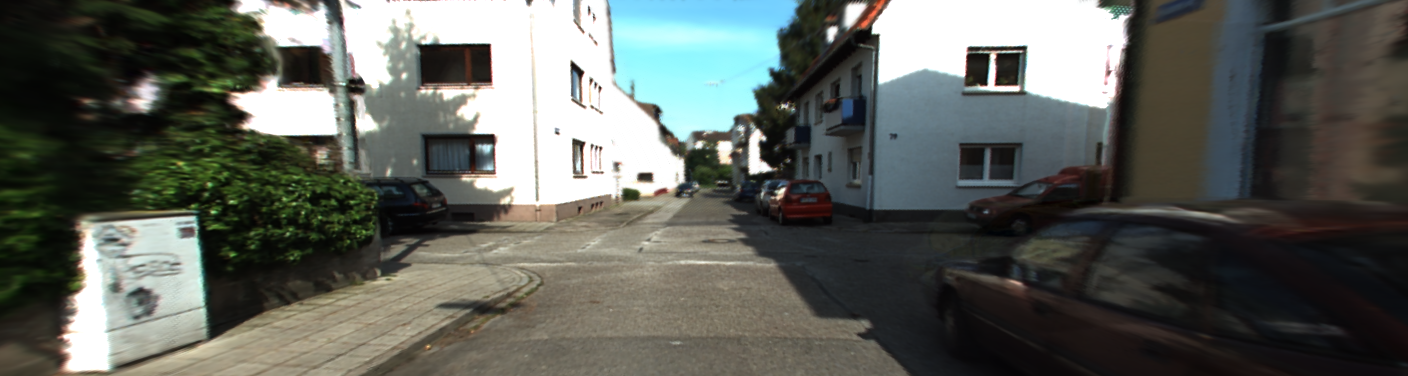} &
      \includegraphics[width=\mywidth\linewidth]{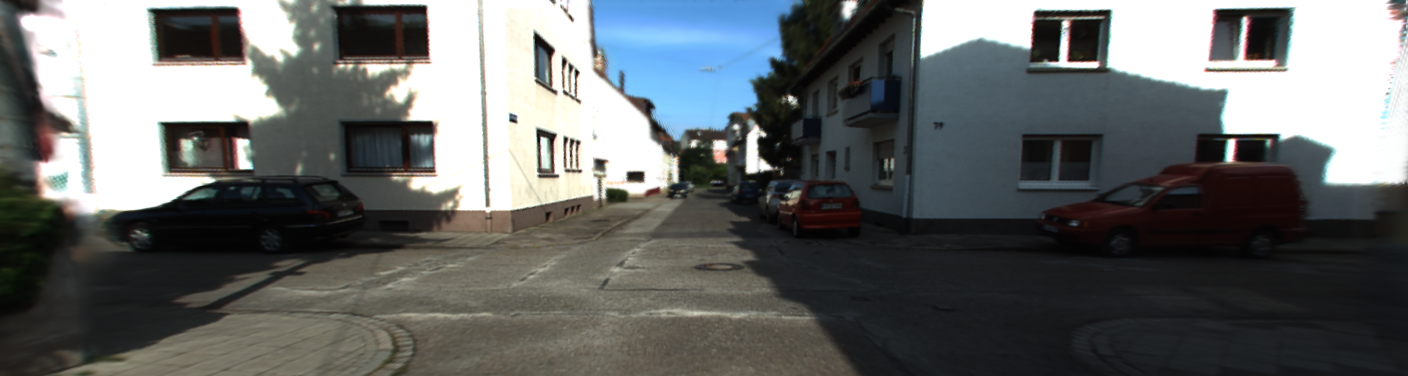} 
      \vspace{\reduceheight}\\
      \includegraphics[width=\mywidth\linewidth]{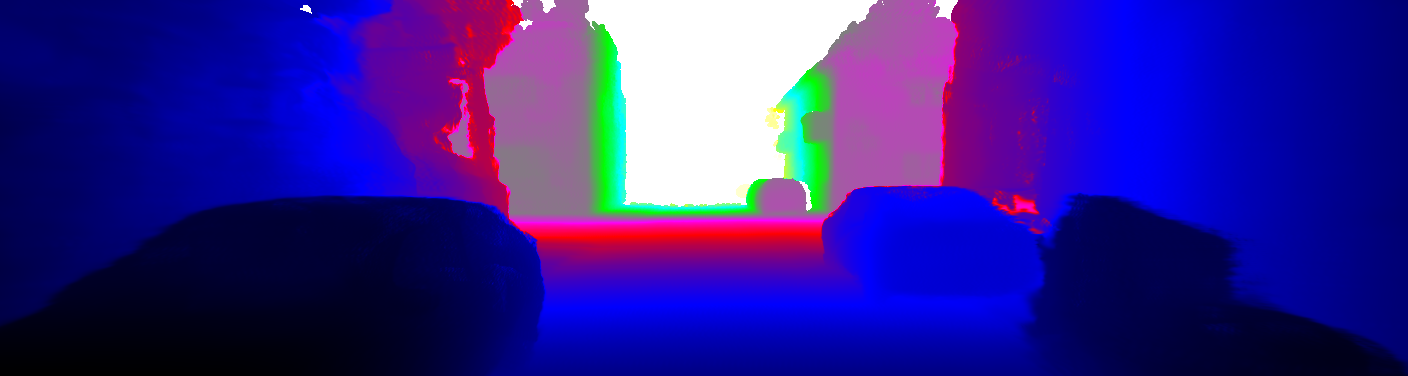} &
      \includegraphics[width=\mywidth\linewidth]{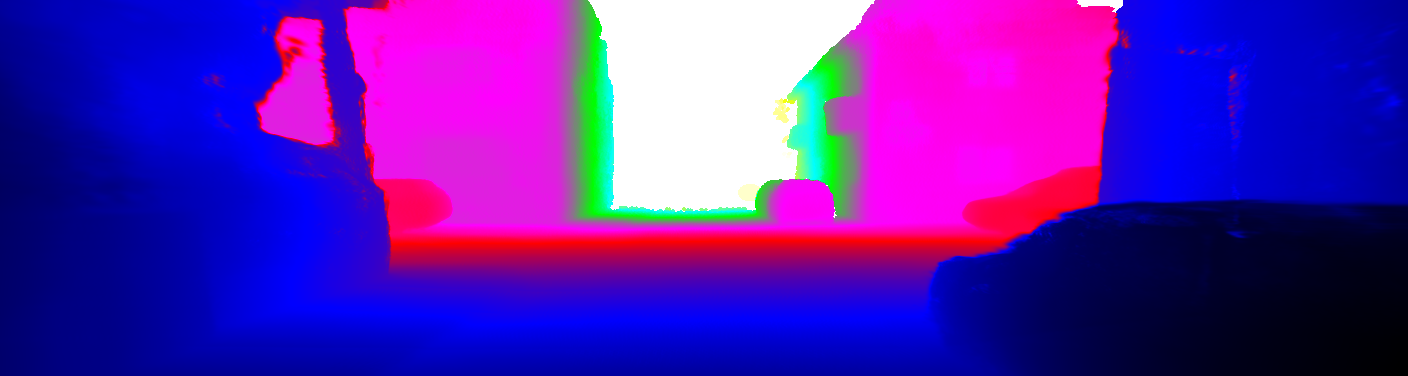} &
      \includegraphics[width=\mywidth\linewidth]{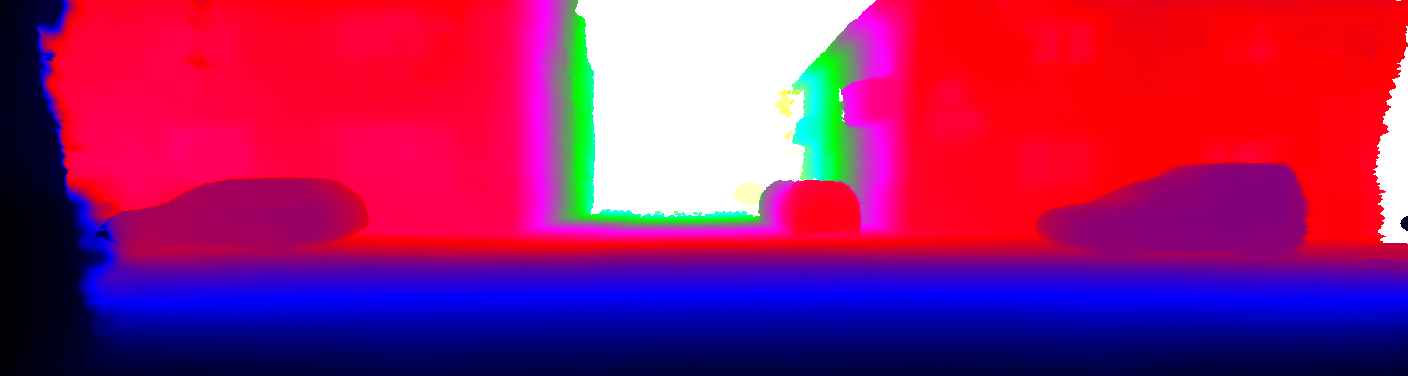} 
      \vspace{\reduceheight}\\
      
     \end{tabular}\vspace{-0.2cm}
     \caption{{\bf More Qualitative Results on KITTI-360}. We visualize synthesized images (odd rows) and the corresponding proxy geometry (even rows) on novel scenes generated by our pretrained model through a feed-forward inference.}
     \label{fig:more res}
     \vspace{-0.021cm}
    \end{figure*}

We present some failure cases in \cref{fig:failure case}. A key limitation of the proposed approach is its dependence on the metric depth estimation. Our method suffers degeneration when the depth model struggles to provide fine-grained depth estimates for thin structures.

Another limitation is that our generalizable hemisphere background only roughly approximates the geometry of distant landscapes, leading to artifacts on background region. However, high-quality rendering of the foreground is generally more critical for autonomous driving applications. A potential solution may be to leverage image-based rendering techniques to model the background, though this would reduce rendering efficiency.

\end{document}